\documentclass{article}

\IfFileExists{iclr2025_conference.sty}{%
  \usepackage{iclr2025_conference,times}%
}{%
  \usepackage[margin=1in]{geometry}%
  \usepackage{times}%
  \usepackage{natbib}%
}
\usepackage{amsmath,amssymb}
\usepackage{graphicx}
\usepackage{booktabs}
\usepackage{url}
\usepackage{tikz}
\usetikzlibrary{arrows.meta,positioning,fit,backgrounds,calc}
\usepackage{hyperref}
\definecolor{maia}{RGB}{31,119,180}
\definecolor{tok}{RGB}{44,160,44}
\definecolor{tx}{RGB}{214,39,40}
\definecolor{outc}{RGB}{148,103,189}
\definecolor{pers}{RGB}{255,127,14}
\definecolor{sf}{RGB}{140,86,75}
\usepackage{pifont}

\IfFileExists{iclr2025_conference.sty}{\iclrfinalcopy}{}  

\title{Matilda: Engine-Agnostic Search with\\ Human Policy Guidance}

\author{Jason Carlson \\
\texttt{jcarlson212@gmail.com}}

\usepackage{xcolor}
\usepackage{placeins}

\begin{document}
\maketitle


\newcommand{\suppref}[2]{Appendix~\ref{#1}}
\newif\ifanonsubmission
\anonsubmissionfalse
\newif\ifstandalonesupplement
\standalonesupplementfalse
%

\begin{abstract} 
Chess engines have evolved from search-based systems optimized for strength to neural 
policies optimized for predicting human decisions. Existing approaches largely separate these goals: 
search engines achieve superhuman strength but poorly model humans, while models such 
as Maia-3 capture rating-conditioned behavior yet degrade at elite levels. We present
\textsc{Matilda}, a modular residual re-ranking architecture that decouples behavioral priors 
from tactical search, combining a frozen human policy with an engine-agnostic search 
backend through a lightweight residual model. \textsc{Matilda} learns residual corrections over 
the full legal-move distribution from frozen policy context, time control, player-style 
embeddings, and search-derived candidate features. A zero-initialized residual head 
exactly recovers the frozen policy before training while optimization minimizes negative 
log-likelihood (NLL). Instantiated with Maia-3 and Stockfish, \textsc{Matilda} reduces human-move 
prediction NLL by $18.5\%$ and raises top-1 accuracy from $60.1\%$ to $66.1\%$ on temporally 
held-out verified-human $3000+$ Elo Lichess blitz games, with player-style embeddings contributing 
a further $1.8\%$ and $+0.2$ percentage points (pp) respectively. Seed-paired ablations attribute 
these gains to search rather than additional data; the findings are replicated in Go -- decomposing expert play 
into recognition and verified calculation. Below $2500$ Elo,
where search annotations are unavailable, \textsc{Matilda} preserves Maia-3's performance.
\end{abstract}

\section{Introduction}
Computer chess has developed through three threads, each with a
different relationship between search and learned policies. Deep Blue
reached superhuman strength through massively parallel alpha-beta search
over a hand-tuned evaluation \citep{campbell2002deepblue} with search as
the sole source of strength. A second thread shifted the burden onto
learned policies that guide or even replace search: AlphaGo
\citep{masteringgo}, AlphaZero \citep{silver2017alphazero}, NNUE within
Stockfish \citep{stockfish}, and pure transformer engines
\citep{monroe2024transformer,ruoss2024grandmaster}. The third thread
changed the objective from the best move to the human one -- the Maia
family \citep{mcilroyyoung2020maia,tang2024maia2,maia3chessformer}
predicts moves conditioned on a skill band, Maia-3 being the most
accurate, with follow-on work modeling individual play
\citep{mcilroyyoung2022individual,mcilroyyoung2022stylometry,tang2025maia4all}
-- Maia makes search disappear entirely, in favor of a direct behavioral
prior.

A fourth direction emerges when the goal is to combine a human
behavioral prior with an independent search process. Allie
\citep{zhang2025allie} explores this in chess by applying search over a
human-aligned policy, but search remains coupled to the prior it
searches through. In Go, a similar approach has been taken: KataGo can
blend its human-imitation network into its superhuman search to play in
a human style \citep{katagohumansl,goprior}. In both types of games the search 
is directly coupled to the prior. Stepping back, this raises two questions with a
long history in the study of expertise
\citep{degroot1965thought,chase1973perception}: how much of expert play
is pattern recognition, and how much is calculation -- and can the two
be modeled as separate, recombinable components? The classical answer
leans toward recognition: de Groot found that masters calculate little
more than weaker players, and Chase and Simon located expert
superiority in perception -- finding the right moves to consider at
all. Yet search still matters at the top, and de Groot's own four-phase
description of expert thought -- orientation, exploration,
investigation, proof -- suggests how the components fit together.

\textsc{Matilda} was ultimately inspired by Botvinnik's idea \citep{botvinnik1970} of creating computer models 
of chess that mimicked the actual cognitive processes grandmasters like him employed; in particular, 
identifying promising zones and searching over their associated trajectories. 
\textsc{Matilda} implements this by -- rather than
distributing computation across every legal move -- first
using a frozen, rating-conditioned Maia-3 to identify a human-plausible
candidate set (orientation and exploration). Search is then restricted
to this subset (investigation), where an external engine (Stockfish, Lc0)
provides engine evaluations to a lightweight Set Transformer
\citep{lee2019settransformer} to re-rank the candidates (proof). This residual 
design allows the re-ranker to learn engine-informed corrections while a 
compact 32-dimensional player embedding captures stylistic deviations, a feature that would 
be far more difficult if search were coupled with the prior. Since 
rating-dependent behavior is already modeled by the frozen base network, the 
style embedding is encouraged to encode player-specific preferences 
rather than playing strength, which we evaluate directly.

\section{Contributions}
\begin{itemize}\itemsep1pt
\item \textbf{\textsc{Matilda}, a modular architecture.} 
A frozen human policy proposes candidate actions, while an independent 
search backend provides tactical evidence through a search-engine-agnostic 
candidate interface. A lightweight permutation-invariant residual model 
reconciles the two, allowing different search engines to be substituted
without modifying the learned architecture, and a compact $32$-d style
vector to personalize the re-ranking without entangling identity with
strength.
\item \textbf{Engine features that pay.} On a $2500+$ Elo blitz/rapid human benchmark ($1.56$M held-out decisions
after excluding engine and Terms of Service-flagged accounts), \textsc{Matilda} improves on Maia-3 by $+0.46\%$ at
$2500$--$2600$ but $+4.3\%$ at $2800$--$2900$, $+11.9\%$ at $2900$--$3000$,
and $+21.9\%$ at $3000+$, while preserving Maia-3 at $1000$--$2500$. Seed-paired ablations -- identical data and seeds,
$\pm$engine features -- attribute the $2800+$ improvement to the engine
features: the attribution is sharpest where the prior's top choice is
tactically refuted and the human avoids it ($+41\%$), which indicates human calculation, not memorized engine
play. We observe the same qualitative behavior in Go, suggesting that the benefit 
arises from the architectural decomposition rather than chess-specific design choices.
\item \textbf{An extensible UCI engine.} We release \textsc{Matilda} as a
UCI (Universal Chess Interface)-compatible engine with a modular search interface, allowing researchers
to replace the search backend with any UCI-compliant engine and experiment
with alternative search strategies without modifying the learned model.
\end{itemize}

\section{Related Work}
\subsection{Human-Like Neural Network Policies.} Maia
\citep{mcilroyyoung2020maia} trains separate networks per rating band to
predict human moves; Maia-2 \citep{tang2024maia2} unifies skill levels with
a skill-aware attention mechanism, and Maia-3 \citep{maia3chessformer}, the
Chessformer architecture, is the current most accurate human move predictor
and our policy's prior. Chessformer's geometric attention bias (GAB) also
yields a per-square importance map -- a signal for where on the board a
player's attention concentrates -- which \textsc{Matilda} consumes as part
of its context encoding. These models capture the population at a
skill level; \textsc{Matilda} models the deviations of strong play and of
individuals from that prior. KataGo's human-imitation network
\citep{katagohumansl} plays the analogous role in Go -- strength and
era-conditioned human move prediction -- and is the prior for our Go
replication.

\subsection{Individual Behavior and Stylometry.}
\citet{mcilroyyoung2022individual} fine-tune Maia per player ($4$--$5$
percentage points of move-matching, requiring $\geq$$5{,}000$ games per player; at $\sim$$1{,}000$ games it is ineffective, with base Maia often
better), and behavioral stylometry \citep{mcilroyyoung2022stylometry}
learns transformer embeddings that re-identify a player from their games
with high accuracy. Skill-group $n$-gram models \citep{zhong2025ngram}
predict moves by rating band. Closest to our style vector, Maia4All
\citep{tang2025maia4all} initializes per-player embeddings from prototype
players, reaching individual move prediction from only $\sim$20 games. These works
recover identity or skill but do not target a style
representation disentangled from strength; our
residual-on-a-rating-conditioned-base design makes Elo-disentanglement an
architectural property and measures it directly, and operates at $2500+$ Elo,
above the club-level range these systems target.

\subsection{Search Engines} AlphaZero \citep{silver2017alphazero}, Deep Blue
\citep{campbell2002deepblue}, and Stockfish \citep{stockfish} define
strong play. We use engines only as sources of per-candidate evaluation
features, drawn deliberately from two families: Stockfish couples deep
alpha--beta search with an incrementally updated lightweight neural evaluation (NNUE);
Lc0 \citep{lc0} is an open-source AlphaZero descendant -- Monte Carlo
tree search guided by a policy/value network under a node budget. The
families differ in how candidates are explored and how positions are
valued (a material-anchored evaluation vs.\ a learned win probability),
yet both -- like nearly every modern engine -- report analysis through the
UCI, whose standard centipawn field
is exactly what our candidate features (score, loss-vs-best, rank) are
computed from. Agreement between the seed-paired Stockfish and Lc0 runs
is therefore evidence that \textsc{Matilda} consumes protocol-level search
evidence any UCI engine can supply, rather than being coupled to one
engine. In Go the same role is played by KataGo \citep{goprior}, consumed
through its analysis interface rather than UCI.

\subsection{Human-Aligned Search} Allie \citep{zhang2025allie} trains a
decoder-only transformer on human games with three heads -- policy, human
pondering time, and value -- under the joint objective
\[
\begin{aligned}
\mathcal{L}(\theta)=\sum_i \Big(&-\log p_\theta(m_i \mid \mathbf{m}_{<i})\\
&+\big(t_\theta(\mathbf{m}_{<i})-t_i\big)^2
+\big(v_\theta(\mathbf{m}_{<i})-v\big)^2\Big),
\end{aligned}
\]
and deploys them inside Monte Carlo Tree Search ({MCTS}) whose implementation follows AlphaZero
\citep{silver2017alphazero}: rollouts pick moves by the {PUCT} rule
\[
Q(s,a)+c\,P(s,a)\,\frac{\sqrt{\sum_b N(s,b)}}{1+N(s,a)},
\]
where the policy head supplies the prior $P(s,a)$, the value head scores
the leaves averaged into $Q(s,a)$, $N$ counts visits, and the rollout
budget $N_{\mathrm{sim}}=\lfloor c'\,t_\theta(\mathbf{m})\rfloor$ is
set by the predicted pondering time. Writing $P_H, V_H$ for these
human-trained policy and value models and $V_E$ for an engine value, the
contrast with our approach is
\[
\begin{array}{ll}
\text{Allie:} & \arg\max\nolimits_a\, \mathrm{MCTS}_{N_{\mathrm{sim}}}(P_H, V_H)\\[2pt]
\textsc{Matilda}: & f\!\left(P_H, \mathrm{search}(V_E)\right)
\end{array}
\]
Allie's search runs on top of human models to choose a human-like
move; \textsc{Matilda} inverts the information flow, learning a re-ranking
of the human prior with engine-search results as input evidence.

\section{Method}

\begin{figure*}[t]
\centering
\includegraphics[width=0.92\textwidth]{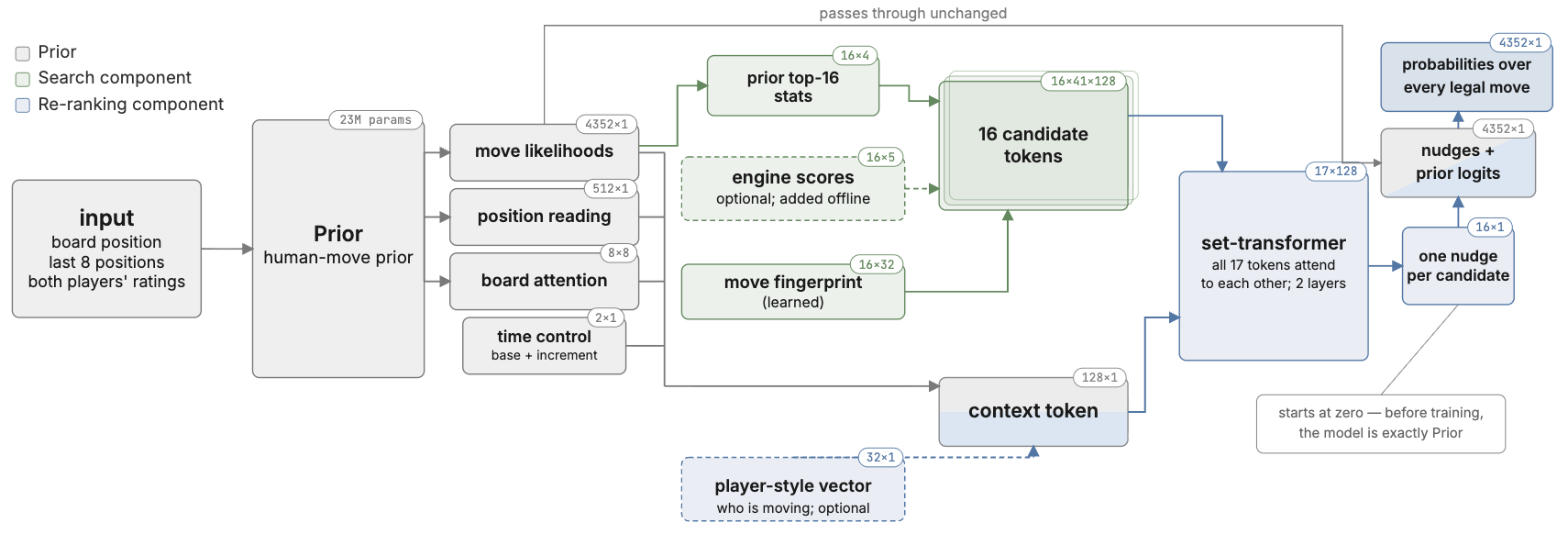}
\caption{\textsc{Matilda}'s architecture, read left to right. The boxed
numbers on each block are the dimensions of the tensors passed along;
the frozen human-move prior is Maia-3 23M. Colors mark the three
components -- gray the frozen prior, green the search evidence, blue the
re-ranker -- and dashed outlines the optional inputs (engine scores,
player-style vector). The Method section walks the figure in full.}
\label{fig:v2architecture}
\end{figure*}

\subsection{Engine-Agnostic Re-Ranking}

\textsc{Matilda} implements the re-ranking decomposition of the
introduction for personalized chess
(Figure~\ref{fig:v2architecture}) by treating
Maia-3's top-$16$ moves as a set of candidate tokens. Each token contains
(i) statistics from the human prior---normalized Maia-3 logit, rank,
gap-to-best, and a top-choice flag; (ii) optional search features---a scored
flag, normalized centipawn evaluation, centipawn loss versus the engine's best
move, engine rank, and an engine-top-choice flag; and (iii) a learned
$32$-d move fingerprint. Two transformer layers ($d{=}128$, $8$ heads, no
positional encoding; $1.7$M parameters, trained in stages on progressively
deeper rating tranches---\suppref{app:training}{D}) attend jointly over the
candidate set, allowing the model to compare candidates directly while
remaining permutation invariant.

Search features are provided on a per-position basis. When available, every
candidate receives search statistics from a single search procedure; otherwise,
the search-feature is set to zero and the scored flag indicates that no search occurred. 
Because the interface consists only of
generic candidate-level summaries (evaluation, loss, rank, and best-move
indicator), all derivable from the standard {UCI} protocol, the same checkpoint operates with or without search
features and accepts outputs from arbitrary search
procedures---Stockfish, Lc0, shallow search, or future
alternatives---without retraining. The number of top candidates we provided 
is $16$ because the played move appears within Maia-3's top-$16$ candidates
on $99.9\%$ of $2500$--$2800$ decisions and $99.7\%$ even at $3000+$
(see \suppref{app:audits}{C} for details).

More formally, \textsc{Matilda} implements
\[
p(\cdot\mid s)
\;=\;
f_\theta\!\big(P_H(s),\,\mathrm{search}(V_E)\big),
\]
where $P_H$ is a frozen human prior and $\mathrm{search}(V_E)$ denotes
an auxiliary computation that provides candidate-level evidence. The
prior maps a position $s$ to
\[
P_H(s)=(\ell,\,h,\,\mathcal{C}),
\]
consisting of full-vocabulary logits
$\ell\in\mathbb{R}^{4352}$, an internal representation $h$ (pooled hidden
state and board attention), and the candidate set
$\mathcal{C}$---Maia-3's top-$16$ moves, with vocabulary indices
$\mathrm{idx}$. The auxiliary computation contributes only
per-candidate summaries,
\[
e_i
=
\mathrm{search}(V_E)(s,c_i)
\in
\mathbb{R}^{5},
\qquad
c_i\in\mathcal{C},
\]
with $e_i\equiv0$ when no search information is available. These two
sources are encoded into one context token and one token per candidate,
\[
z_0=\mathrm{enc}_0(\ell,h,\mathrm{tc},u),
\quad
z_i=\mathrm{enc}_c\!\big(\mathrm{stats}_i(\ell),e_i,\mathrm{emb}(c_i)\big),
\]
where $\mathrm{stats}_i(\ell)$ are candidate $i$'s prior statistics
(item i above), $\mathrm{emb}(c_i)$ its move fingerprint (item iii),
$\mathrm{tc}$ the time control, and $u$ the optional style vector. The
set transformer $g_\theta$ -- the two layers above -- ends in a
zero-initialized linear head predicting one logit adjustment per
candidate,
\[
\delta
=
g_\theta(z_0,\dots,z_{16})
\in
\mathbb{R}^{16},
\qquad
\tilde{\ell}
=
\ell
+
\mathrm{scatter}(\delta,\mathrm{idx}),
\]
where $\mathrm{scatter}(\delta,\mathrm{idx})\in\mathbb{R}^{4352}$
carries $\delta_i$ at vocabulary index $\mathrm{idx}_i$ (candidate
$i$'s move) and zero at every other index -- only the $16$ candidates'
logits move, each by its own adjustment. The adjusted logits are then
normalized over the legal-move set $\mathcal{A}(s)$:
\[
p(a\mid s)
=
\mathrm{softmax}\!\left(
\tilde{\ell}
+
\log\mathbf{1}[a\in\mathcal{A}(s)]
\right)_a.
\]

This factorization makes the roles of the two inputs explicit:
$P_H(s)$ contributes the prior's representation and candidate set,
whereas $\mathrm{search}(V_E)$ contributes only the candidate evidence
$e_{1:16}$. Setting $e\equiv0$ recovers the search-free
regime within the same parameters. Because the output head is zero-initialized, 
the untrained
model reproduces Maia-3 logit-for-logit, so every improvement is
value-added. Moves outside the top-$16$ keep their logits, their
probabilities changing only through the shared normalization.

\subsection{Personalization: Player Embedding Design}
Identity is one $32$-d embedding row per player, passed through a
zero-initialized projection and added to the context token (dashed blue
in Figure~\ref{fig:v2architecture}); During training the
true id is replaced by the reserved generic row $0$ with probability
$0.1$ (embedding dropout). The embedding table and its projection use 
only $0.2$M parameters with the rest of \textsc{Matilda} frozen during training.
New players are fitted post-hoc: only their embedding rows train, with
\textsc{Matilda} and the shared projection fixed.

\subsection{Context Design}
Candidates are judged against one context token that carries everything
global (Figure~\ref{fig:v2architecture}). Frozen Maia-3 --
which reads the position, the last eight positions, and both players'
ratings -- contributes three signals through
small learned encoders: its full policy logits (compressed
$4352{\to}256{\to}128$), its pooled hidden state ($\mathbb{R}^{512}$,
L2-normalized), and its $8{\times}8$ board-attention map. Throughout we 
use Maia-3 23M instead of the Maia-3 73M parameter variant since it's both 
faster to run and obtains over 
$99\%$ of the overall accuracy of the larger model. Two
inputs bypass Maia-3 directly to the context token's concatenation: the
time-control pair $(\log(1{+}b),\log(1{+}i))$, base seconds $b$ and
increment $i$ and the optional style
vector. The concatenation projects to one $128$-d token.

\subsection{Data}
All data is drawn from the public Lichess monthly dumps
\citep{lichessdb}, rated Blitz and Rapid,
over a 12-month window (2025-06 to 2026-05); the $3000+$ tranche instead
draws on 24 earlier months (2023-01 to 2025-05), disjoint from this window,
to offset elite-play scarcity. A decision is one moved position from ply $\geq2$ 
to account for purely memorized opening prep in ply $1$;
Table~\ref{tab:tranches} lists the tranches. The base re-ranker trains
on the pool, the pool remainder, the GM tranche, and the train split of
the anonymous low-Elo rows (the eval split is the low-Elo preservation
benchmark); \textsc{Matilda} continues from it on the $3000+$ tranche,
keeping every tenth shard of that original mix in the training stream so the
$3000+$ Elo elite specialization does not erase the broader model. Style embeddings and their shared
projection are then trained on pool and GM rows with \textsc{Matilda}
frozen. The club tranche is never trained on: it only fits new embedding
rows post-hoc, testing personalization below the pool's rating range. Every
selection is seed-deterministic; every split is temporal per player
(first $85\%$ train, last $15\%$ evaluation-only), with $10\%$ of players
held out entirely and exact (player, position, move) deduplication against
all evaluation splits before any row enters training. Each decision is
featurized once with frozen Maia-3; Stockfish annotations are produced on
the candidate set only (\texttt{searchmoves} restricted to the $16$
candidates, \texttt{multipv}$=$$16$, depth $14$/$10$s) for all $2800+$
rows, $2500+$ rapid rows, and the $3000+$ tranche -- $10.2$M search annotated
rows total, with a $181$k-row sample re-searched at depth $21$/$30$s ($89{,}692$
verified-human rows) showing depth insensitivity (\suppref{app:audits}{C}).

\begin{table}[t]
\centering
{\small
\begin{tabular*}{\columnwidth}{@{\extracolsep{\fill}}llr@{}}
\toprule
tranche & filter & train/eval/holdout \\
\midrule
pool & both $\geq2500$ & $6.9$M/$1.2$M/$0.9$M \\
pool remainder & not selected & $4.9$M/--/-- \\
GM & mover $\geq2800$ & $4.5$M/$0.8$M/$0.5$M \\
$3000+$ (24\,mo) & mover $\geq3000$ & $2.0$M/$0.35$M/$0.24$M \\
club & mover $1000$--$2500$ & $4.9$M/$0.86$M/$0.71$M \\
anon.\ low-Elo & $1000$--$2500$ & $1.7$M/$0.43$M/-- \\
\bottomrule
\end{tabular*}}
\caption{Data tranches, all rated Blitz and Rapid (train/eval/holdout
decisions). Named-player tranches require minimum per-player history and
cap per-player contributions; the bot/TOS exclusion (below) is applied in
the data loaders before all training and evaluation. Full selection
rules, caps, and player counts: \suppref{app:audits}{C}.}
\label{tab:tranches}
\end{table}

\textbf{Time controls.} Rapid is $\approx$$6\%$ of the $2500+$ pool, and
each rapid decision is drawn four times per epoch in training (a plain
$4\times$ row repetition; \suppref{app:training}{D}); gains
are reported per time class
(blitz-trained Maia-3 has higher negative
log-likelihood, NLL, on rapid rows). All reported gains are
matched-row deltas (baseline and model scored on identical decisions), and
candidates are Maia-3's own top-$16$ with the played move never injected
(see \suppref{app:audits}{C} for audits).

\textbf{Account quality (the bot audit).} Rated play filters game
seriousness, not account type: Lichess BOT-titled engine accounts and
Terms Of Service-flagged (TOS) accounts dominate the public high-rating
population -- $12$--$42\%$ of unfiltered $2800$--$2900$ extractions,
$59$--$77\%$ at $2900$--$3000$, and $76$--$98\%$ of the $3000+$ tranches, while the club
range is essentially clean. We resolve every account's title via the
Lichess API and exclude BOT-titled and TOS-flagged accounts ($2{,}892$
accounts; $21.5\%$ of raw rows) from both training and evaluation,
enforced by per-file exclusion masks and an automated conformance suite. All numbers
in this paper are on the human benchmark.

\section{Experiments}
All results use full-vocabulary NLL on temporally held-out decisions of
verified-human accounts, with $95\%$ player-stratified bootstrap CIs
\citep{agarwal2021deep} (the player set resampled with replacement $2{,}000$
times). Relative gain
is
\[100\,(\mathrm{NLL}_{\text{Maia-3}}-\mathrm{NLL}_{\text{model}})/\mathrm{NLL}_{\text{Maia-3}}.\]
In the technical supplement: audits and reproduction commands in \suppref{app:audits}{C}, 
qualitative breakdowns and famous-player case studies in \suppref{app:quali}{B}, 
training staging in \suppref{app:training}{D}, the checkpoint inventory in \suppref{app:checkpoints}{A}.

\begin{table*}[t]
\centering
\begin{tabular}{lrcccc}
\toprule
Elo band & $n$ & \textsc{Matilda} vs Maia-3 & $+$personal vs Maia-3 & identity & top-1 \% (M3$\to$$+$p) \\
\midrule
2500--2600 & 723,280 & $+0.46$ [0.42,0.51] & $+0.76$ [0.70,0.81] & $+0.29$ [0.26,0.33] & 62.4$\to$62.5 \\
2600--2700 & 283,467 & $+0.58$ [0.50,0.68] & $+0.88$ [0.77,0.98] & $+0.29$ [0.23,0.34] & 62.8$\to$63.0 \\
2700--2800 & 113,470 & $+0.56$ [0.42,0.70] & $+0.81$ [0.66,0.99] & $+0.25$ [0.17,0.33] & 62.9$\to$63.1 \\
2800--2900 & 313,492 & $+4.26$ [3.87,4.77] & $+4.70$ [4.25,5.32] & $+0.47$ [0.35,0.59] & 60.8$\to$62.3 \\
2900--3000 & 89,426 & $+11.95$ [9.75,14.48] & $+13.26$ [10.79,16.06] & $+1.46$ [1.07,1.93] & 59.8$\to$64.1 \\
3000--4000 & 36,650 & $+21.91$ [17.80,26.05] & $+22.85$ [18.87,26.92] & $+1.12$ [0.46,1.79] & 61.1$\to$68.4 \\
\midrule
2500+ overall & 1,559,785 & $+2.53$ [2.23,2.84] & $+2.93$ [2.62,3.27] & $+0.41$ [0.36,0.45] & 62.0$\to$62.8 \\
\bottomrule
\end{tabular}

\caption{\textsc{Matilda} with and without style embeddings vs.\ Maia-3 23M
(the variant used throughout) by Elo band: relative NLL improvement over all legal moves, $95\%$
player-stratified bootstrap CIs; the last column is absolute top-1
accuracy from M3 (Maia-3) to $+$p (\textsc{Matilda} with the personal
style vector). Eval is from temporally held-out decisions of
verified-human accounts; Maia-3 is blitz-trained, so
Table~\ref{tab:blitzonly} gives the like-for-like blitz-only slice and the
no-engine variant (Stockfish-ablation section) bounds the
training-distribution confound.}
\label{tab:mainbands}
\end{table*}

\subsection{Engine Features Where the Prior Runs Out}
Table~\ref{tab:mainbands} is the central result. The pattern matches the
thesis's prediction: sub-$1\%$ gains where Maia-3 is in-domain
($+0.46$--$0.58\%$ at $2500$--$2800$), then $+4.3\%$ at $2800$--$2900$,
$+11.9\%$ at $2900$--$3000$, and $+21.9\%$ at $3000+$, where Maia-3's
rating conditioning loses its grounding and calculation becomes the
differentiating factor. The $3000+$ band is
small in absolute terms -- $105$ verified-human elite players -- which the
player-stratified CIs make clear, and its gain is broad-based rather
than driven by prolific accounts: $91\%$ of those players individually
improve. One asymmetry worth mentioning: at
$2900$--$3000$ the typical-player gain is smaller than the row-weighted one
(interquartile mean $+7.0\%$ vs $+11.9\%$) -- high-volume players gain more -- so we
report row-weighted numbers as the per-decision quantity, with
player-resampled intervals and per-player distributions in the technical
supplement. Because Maia-3 is blitz-trained and
the benchmark mixes blitz and rapid, we also report the like-for-like
blitz-only slice (Table~\ref{tab:blitzonly}): the top-end result stands
essentially unchanged ($+18.5\%$ vs.\ $+21.9\%$ mixed at $3000+$), while below
$2800$ the mixed aggregate overstates the gain -- so we quote
blitz-only wherever the claim is ``improves Maia-3.'' Below the benchmark range, on an anonymous $432$k-row
evaluation across $1000$--$2500$ where the model has no engine features by
design, \textsc{Matilda} tracks Maia-3 within $-0.13\%$ to $+0.19\%$
($+0.02\%$ overall). This indicates no low-Elo collapse, so one checkpoint serves the
whole rating ladder.

\begin{table}[t]
\centering
{\small\setlength{\tabcolsep}{2pt}
\begin{tabular*}{\columnwidth}{@{\extracolsep{\fill}}lccc@{}}
\toprule
Elo band & rapid share & blitz (like-for-like) & rapid (ext.) \\
\midrule
2700--2800 & $2.7\%$ & $+0.38\%$ & $+6.78\%$ \\
2800--2900 & $1.5\%$ & $+4.15\%$ & $+11.10\%$ \\
2900--3000 & $3.0\%$ & $+11.60\%$ & $+23.14\%$ \\
3000--4000 & $26.4\%$ & $+18.47\%$ & $+32.30\%$ \\
2500+ overall & $4.3\%$ & $+2.23\%$ & $+9.28\%$ \\
\bottomrule
\end{tabular*}
}
\caption{Like-for-like vs.\ capability extension: \textsc{Matilda} vs.\
Maia-3 by time class (same verified-human rows as Table~\ref{tab:mainbands}).
Maia-3 is blitz-trained; rapid is out of its domain.}
\label{tab:blitzonly}
\end{table}

\subsubsection{The Stockfish ablation.}
\begin{figure}[t]
\centering
\includegraphics[width=0.9\columnwidth]{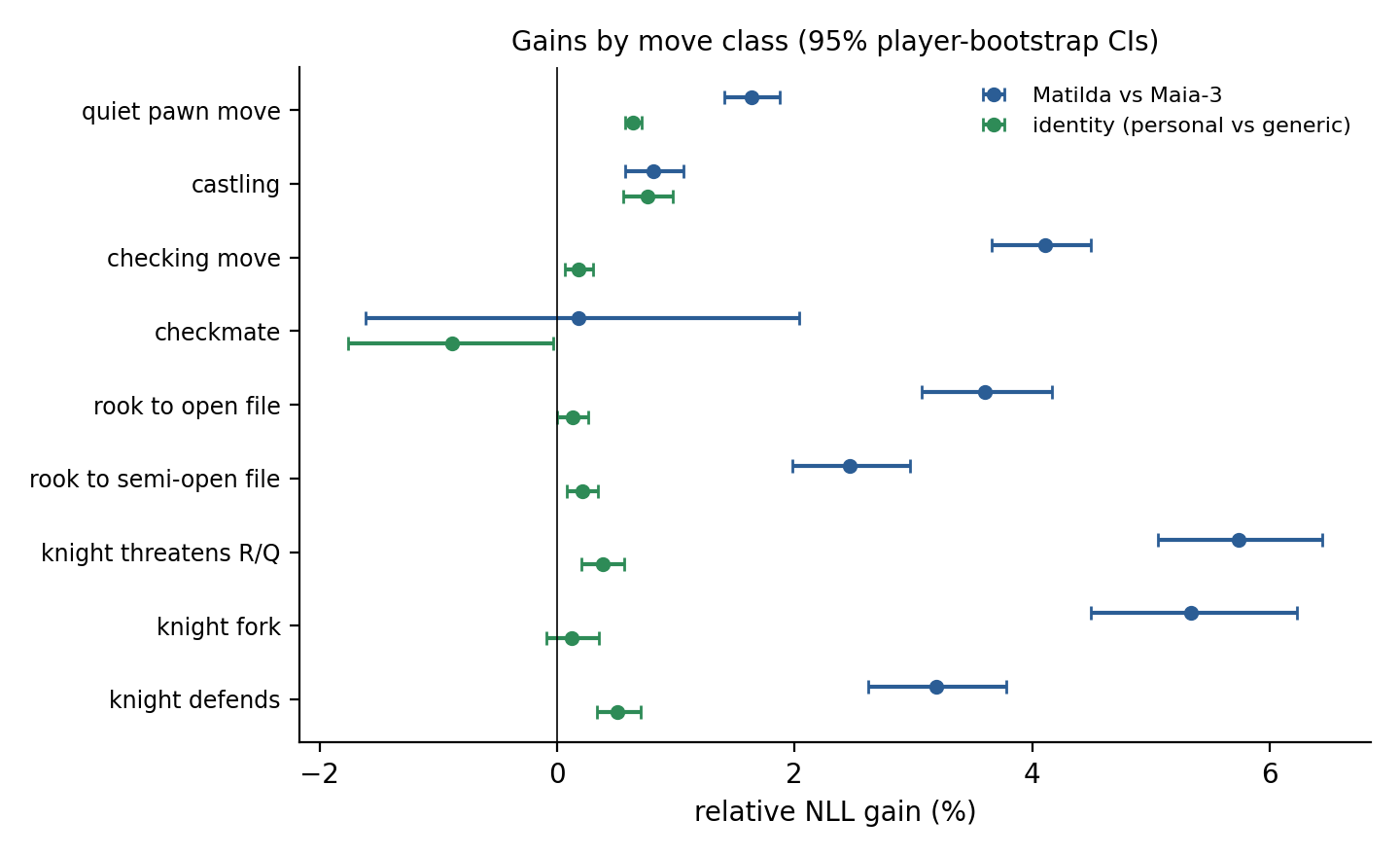}
\caption{Gains by move class. \textsc{Matilda} (blue) is measured
against Maia-3; identity (green) is the personal-versus-generic
embedding delta on the same rows, i.e.\ relative to generic
\textsc{Matilda}. Strongest where tactics are sharpest; counts and
definitions in \suppref{app:quali}{B}.}
\label{fig:movecls}
\end{figure}

Where does the $2800+$ gradient come from? Before committing to
large-scale annotation we trained two variants of the architecture on the
$2800+$ subset -- identical data, seeds, and schedule, differing only in
whether the engine candidate-feature group is present. The difference is
the engine marginal (Table~\ref{tab:engines}, Stockfish column): $+2.3\%$ relative
NLL at $2800$--$2900$, $+5.1\%$ at $2900$--$3000$, $+9.6\%$ at $3000+$.
The pair is trained at reduced scale (its with-engine variant reaches
$+10.1\%$ over Maia-3 at $3000+$), so these are attribution numbers, not
shares of Table~\ref{tab:mainbands}'s gains, which come from the fully
trained model.
The no-engine variant itself gains only $+0.5\%$ over Maia-3 across these
bands ($+0.54$/$+0.39$/$+0.48$; $\pm$engine evaluation rows): the additional
high-Elo training data, time-control conditioning, and re-ranking
architecture do not, by themselves, produce the gradient -- and the same
half percent bounds the training-distribution confound of
Table~\ref{tab:mainbands}. The engine marginal is largest ($+41\%$) where the
prior's own top choice was tactically refuted and the human avoided the
mistake, negative where the human deviated, and concentrated in the
middlegame and endgame rather than memorized openings
(\suppref{app:quali}{B}). Figure~\ref{fig:movecls} shows the same fact
by move class: the gains are strongest on tactically pointed moves --
knight threats, checks, moving a rook to an open file -- and neutral on
checkmates, which the prior already finds.

 The search evidence thus supplies the tactical verification 
the prior lacks: the re-ranker learns to override the prior exactly where its preferred 
move fails calculation and the human did not follow it into the mistake -- a learned
correction, not a memorized preference for engine moves (full
decomposition: \suppref{app:quali}{B}).

\subsubsection{Swapping the engine family: Lc0.}
Is the mechanism Stockfish-specific? We re-annotate the same $2800+$ subset
with Lc0 \citep{lc0} -- an AlphaZero-family engine (MCTS over a
policy/value network, $800$-node budget) rather than alpha--beta over NNUE -- and train a third
seed-paired variant on those features. Table~\ref{tab:engines} reports both
engine marginals side by side: they are essentially identical
through $2800$--$3000$, with Stockfish ahead only at $3000+$ ($+9.6$ vs
$+6.6$), where the fixed $800$-node budget is a far weaker search than
depth-$14$ NNUE. The candidate-feature interface consumes any
strong UCI engine's scores, making \textsc{Matilda} largely engine
agnostic. (Annotation settings and the
depth-insensitivity check are in \suppref{app:audits}{C}.)

\begin{table}[t]
\centering
{\small
\begin{tabular*}{\columnwidth}{@{\extracolsep{\fill}}lrcc@{}}
\toprule
Elo band & $n$ & Stockfish marginal & Lc0 marginal \\
\midrule
2800--2900 & 35,528 & $+2.3\%$ & $+2.5\%$ \\
2900--3000 & 10,018 & $+5.1\%$ & $+5.1\%$ \\
3000--4000 & 3,881 & $+9.6\%$ & $+6.6\%$ \\
2800--4000 & 49,427 & $+3.4\%$ & $+3.3\%$ \\
\bottomrule
\end{tabular*}
}
\caption{Engine-family robustness: the engine marginal (gain over the
no-engine variant) with Stockfish features vs.\ Lc0 features, seed-paired
on identical data (verified-human rows; bot/TOS accounts excluded from
training and evaluation throughout).}
\label{tab:engines}
\end{table}

\subsection{Adding Style Embeddings}
The only trainable parameters are the $32$-d embedding table and its
zero-initialized projection ($0.2$M parameters) on the frozen
\textsc{Matilda}, so the reported delta is the marginal value of
player identity. The pool is $6{,}237$ players at $2500+$ plus the
$1{,}458$-player GM tranche; embedding-dropout of $0.1$ trains id~$0$ as an
explicit generic player, to which unknown players fall back gracefully.
Identity -- personal versus generic vector on the same decisions -- is
worth $+0.41\%$ overall and $+1.1$--$1.5\%$ at grandmaster bands
(Table~\ref{tab:mainbands}).

\begin{figure}[t]
\centering
\includegraphics[width=0.98\columnwidth]{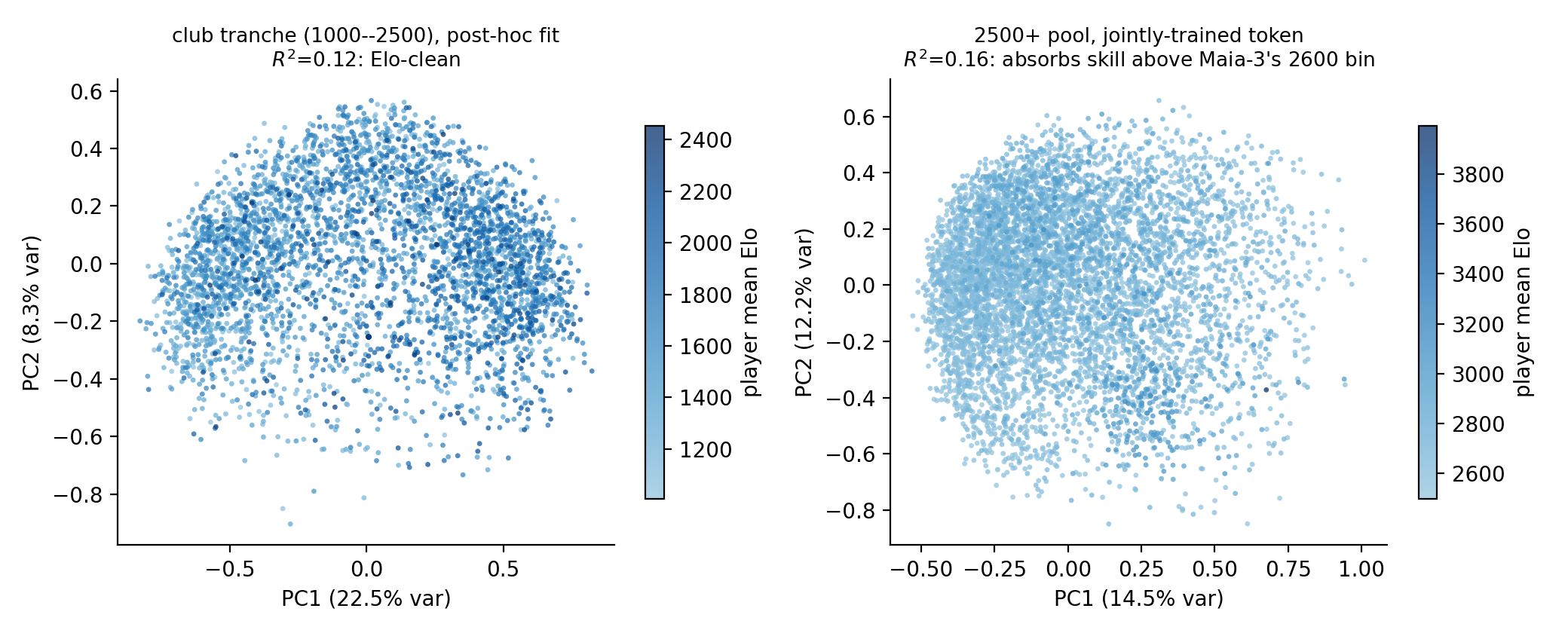}
\caption{Elo-disentanglement, two regimes. Each panel is a PCA of that
population's embeddings in its own basis, colored by player mean Elo.
Left: club-range post-hoc embeddings carry little rating signal
($R^2{=}0.12$). Right: the $2500+$ vector carries somewhat more
($R^2{=}0.16$) -- above Maia-3's effective $2600$ rating ceiling the
embedding is the only channel that can express within-band strength. The two
plotted components capture only a quarter to a third of each population's
embedding variance; the style space is genuinely high-dimensional, with the
full top-$10$ spectrum in \suppref{app:audits}{C}.}
\label{fig:disent}
\end{figure}

\textbf{Where does the added expressivity go?} A cross-validated ridge
probe predicting a player's mean rating from their (L2-normalized) $32$-d
embedding recovers little: $R^2{=}0.12$ for the club-range post-hoc
embeddings and $R^2{=}0.16$ for the jointly-trained $2500+$ vector
(the same probe scores $R^2{\approx}{-}0.01$ on randomly permuted ratings,
i.e.\ chance) -- the latter's mildly higher leakage is
structural, since Maia-3's rating conditioning is unreliable above
$2600$, leaving the embedding as the only channel that can reliably express
within-band strength
(Figure~\ref{fig:disent}; \suppref{app:quali}{B} additionally places six
publicly-attributed grandmaster accounts inside the same space). The
grandmaster-band identity gains therefore
mix style with some within-band skill.

\subsubsection{Personalization at lower ratings.}
\begin{figure}[t]
\centering
\includegraphics[width=0.95\columnwidth]{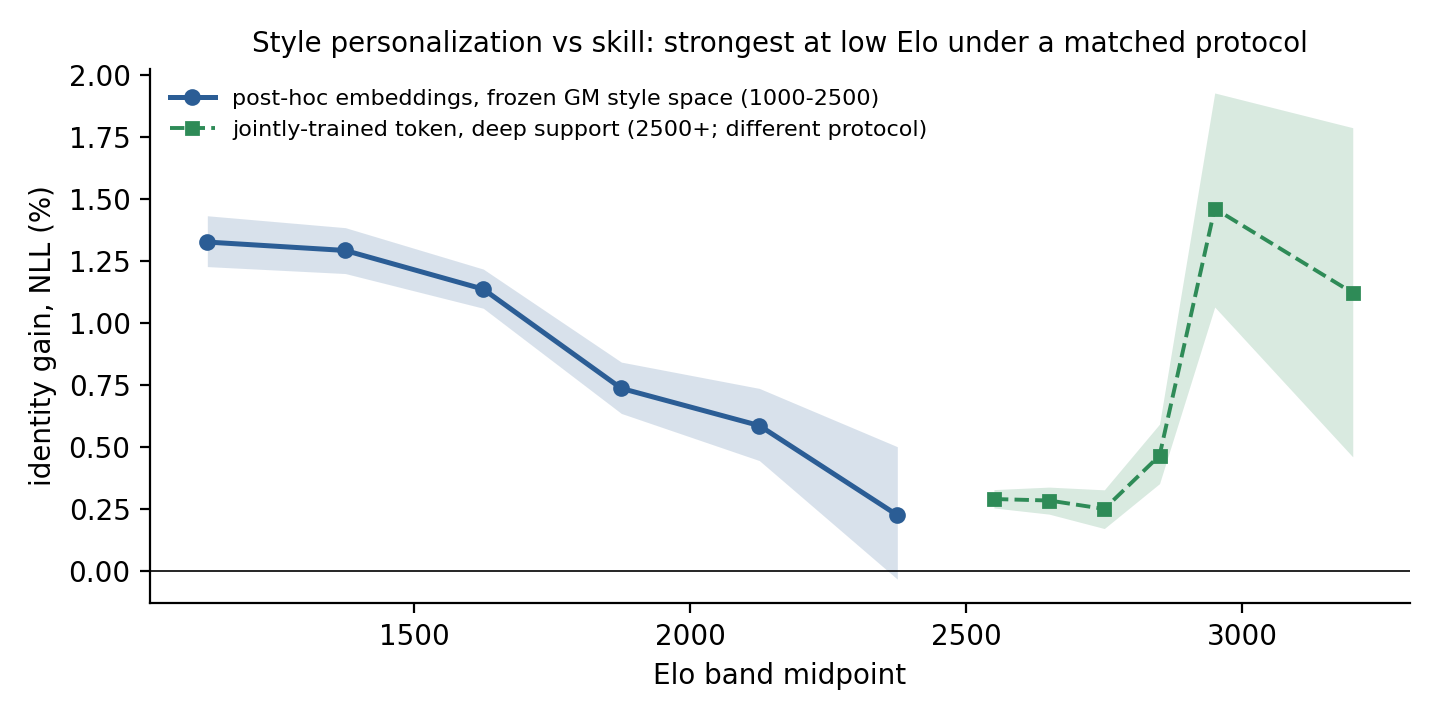}
\caption{Identity gain vs.\ Elo; shaded bands are $95\%$
player-bootstrap CIs. Blue: post-hoc embeddings in the frozen style
space, bands $1000$--$2500$ -- monotone in skill. Green: the
jointly-trained $2500+$ vector (more per-player fitting data, different
protocol), for context.}
\label{fig:identityelo}
\end{figure}

Is style easier to capture where play is less constrained? We freeze
everything -- \textsc{Matilda} and the style projection learned at $2500+$
-- and fit only new $32$-d embedding rows for $3{,}922$ club players
($1000$--$2500$, same temporal split). The style space learned from strong
players transfers, and the identity gain grows monotonically as
skill falls: $+0.28\%$ at $2250$--$2500$ rising to $+1.39\%$ at
$1000$--$1250$, $+1.10\%$ [$1.05,1.15$] overall (Figure~\ref{fig:identityelo}) -- under a matched protocol,
personalization is strongest at the bottom of the ladder, where the space
of reasonable moves is widest.

\textbf{New-player learning curve.} On players held out from all
training, fitting an embedding on a player's first $k$ decisions breaks
even near $k{=}60$ -- personalization pays after roughly two games --
and Maia4All-style prototype initialization removes the cold start at
the price of four times the rating leakage; the full k-shot sweep and
the trade-off are in \suppref{app:quali}{B}.

In absolute terms, at $3000+$ the model moves top-$1$ accuracy
from $61.1\%$ to $68.4\%$ while cutting NLL by nearly a quarter; the
full absolute table across the ladder ($1000$--$4000$), and its
consistency with Maia-3's published move-matching, are in
\suppref{app:quali}{B}.

\subsection{Beyond Chess: Go}
\begin{figure}[t]
\centering
\includegraphics[width=0.92\columnwidth]{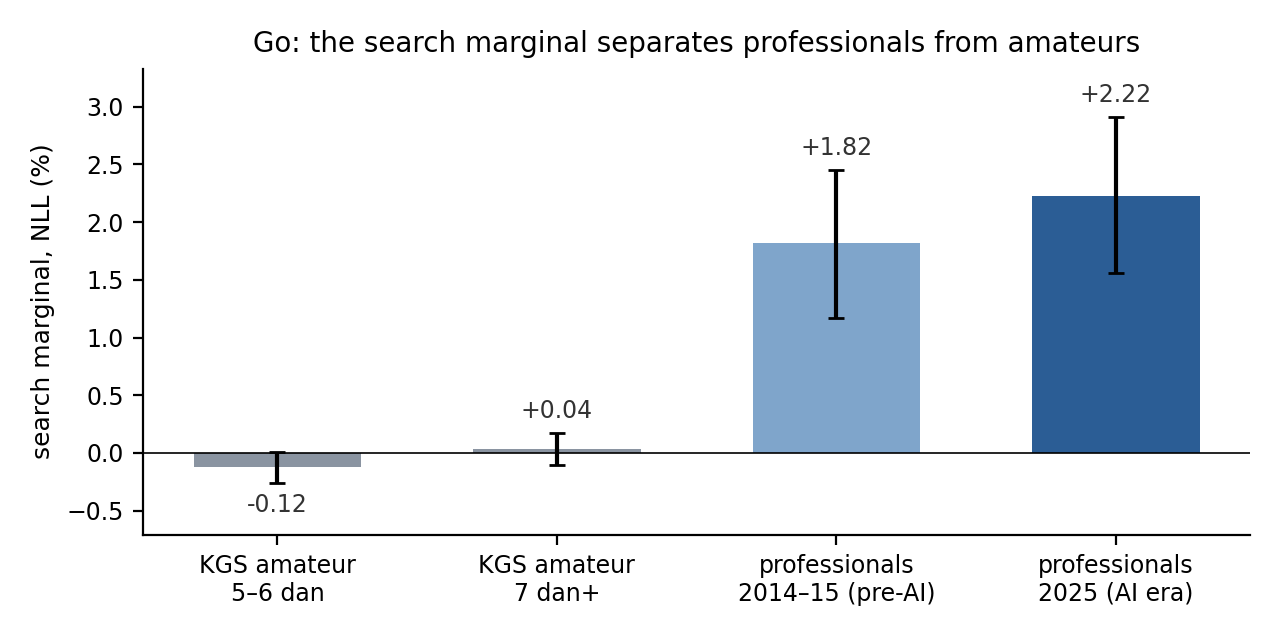}
\caption{The Go generalization: the search marginal (with-search minus
no-search model, seed-paired on identical data; $95\%$ player-bootstrap
CIs) is $\sim$$0$ for both amateur bands but clearly positive for
professionals of both eras (\suppref{app:go}{E}).}
\label{fig:gomarginal}
\end{figure}

We port the architecture unchanged to Go --
KataGo's frozen human-imitation network \citep{katagohumansl} as the
prior, a KataGo search \citep{goprior} over each position's candidates
as the evidence, the same zero-initialized residual re-ranker -- and
train seed-paired $\pm$search models on KGS amateur games from
2005--2015 \citep{ugokgs} and on professional games from 2014--15 and
2025 \citep{gogod,kifubara}. The search marginal is
$\sim$$0$ for amateurs ($-0.12\%$ at $5$--$6$ dan, $+0.04\%$ at $7$
dan$+$) but positive for professionals of both eras: $+1.82\%$
[$1.17,2.45$] for 2014--15 and $+2.22\%$ [$1.56,2.91$] for 2025
(Figure~\ref{fig:gomarginal}). The two professional eras, despite the introduction
of AlphaGo and engine study between them, are
statistically indistinguishable – indicating it's not simply memorized engine moves 
in the post-engine era contributing to the search marginal. Note that the magnitude of improvement in Go is not directly 
comparable to chess since the Go models train on $40$k (professional)
to $206$k--$249$k (amateur) decisions rather than millions, and the Go
prior's era conditioning stays in-domain for professionals where
Maia-3's rating conditioning at $3000+$ does not. The remaining Go data, protocol, 
mechanism slices, and full result specifications are in \suppref{app:go}{E}.

\subsection{Robustness}
\textbf{Memorization.} Temporal-per-player splits let recurring openings
have an exact (player, position, move) key on both sides of the train-eval split
($10.8\%$ of benchmark eval rows; $23\%$ at $3000+$). Excluding every
such row barely moves the headline ($+20.9\%$ at $3000+$ on the unseen-key slice, vs $+21.9\%$ on
all rows). \textbf{Game
identity.} A game-overlap audit re-extracts game ids and removes every eval row
whose game contributed any training row (either side); at $3000+$ the
surviving slice gives $+24.0\%$ [$17.8,30.8$], statistically
indistinguishable from the all-rows $+21.9\%$ [$17.7,26.1$] (only $3{,}510$
rows over $43$ players survive, which the wide interval reflects). A
matched-protocol slice modeled on Maia-3's published evaluation protocol (ply
$\geq$$10$, mover clock $\geq$$30$s) gives $+24.2\%$ [$19.8,28.5$] --
directionally higher, as opening theory is where Maia-3 is relatively
strongest, but again overlapping. The conclusion these audits support is 
that no measurable part of the top-band gain is attributable
to game overlap or to the evaluation protocol. \textbf{Accounts.} BOT/TOS
exclusion is enforced in the data loaders and verified end to end. \textbf{Engine signal.} A dual-depth sample
(the same rows engine-scored at both the production depth $14$/$10$s and a
deeper depth $21$/$30$s; $89{,}692$ verified-human rows) shows the
pipeline is insensitive to annotation depth, $+0.02$pp at $2800$--$3000$,
$+1.9$pp at $3000+$, and the Lc0 swap above shows it is not engine-specific.
Full audit methodology and numbers are in
\suppref{app:audits}{C}; qualitative breakdowns by phase, move class,
and time pressure in \suppref{app:quali}{B}.

\section{Discussion and Conclusion}
We introduced \textsc{Matilda}, a residual re-ranking architecture that
separates three sources of information: a frozen learned prior for
population behavior, an optional compute-intensive search that supplies
per-candidate evidence, and lightweight personalization for individual
deviations.

The chess and Go results together make it clear what the search evidence captures is de Groot's
investigation phase, the deep verification of a few candidate
moves, which a recognition-trained prior cannot supply. The marginal
behaves exactly like the phase should: it pays most where the prior's
preferred move fails tactically and the expert finds the
refutation-proof choice, it appears only for populations whose
calculation outruns what the prior already models ($2800+$ chess and Go
professionals of both eras, but not amateurs), and it predates engines
where pre-AI professionals converged on search-approved moves by
calculation alone. 

The architectural decoupling then gave the other results almost for free: player 
embeddings that capture identity
while strength stays absorbed by the frozen prior, search backends that
swap without retraining (Stockfish to Lc0), and transfer across domains
(chess to Go) by replacing the prior and the engine while keeping the
re-ranker's contract.

More broadly, the results suggest that residual candidate  
re-ranking is a scalable way to combine learned priors with  
specialized computation. The candidate interface isolates the  
expensive reasoning component from the learned policy, al-  
lowing stronger search procedures, alternative engines, or  
future planning algorithms to improve prediction without  
changing the architecture itself. We believe this modularity is  
the primary contribution of Matilda: rather than designing a  
better chess engine, it provides a general framework for inte-  
grating domain-specific computation into learned decision-  
making systems.

\paragraph{Future work.} On the search side, adaptive search
allocation over human-plausible candidates -- extending
Botvinnik's vision \citep{botvinnik1970} toward cheaper computation and
sharper tactics, and deepening the exploration phase
\citep{degroot1965thought} that our fixed candidate set covers only
minimally; on the personalization side, richer style
representations that stay disentangled from strength given thousands of
games per player; and
instantiations wherever experts consult stronger tools (theorem
proving, program synthesis).

\FloatBarrier
\bibliographystyle{plainnat}
\bibliography{references}
\appendix
\section{Models Referenced in This Supplement}
\label{app:checkpoints}
The supplement's experiments compare several checkpoints from
\textsc{Matilda}'s staged construction. All share the identical
architecture (the main paper's Method); they differ only in training data
and starting point. Code names in parentheses match the released
checkpoints. On-disk split names: \texttt{support}/\texttt{query} are the
temporal train/eval splits of each tranche; for the personalization
tranches the support split serves as the embedding \emph{fit} split -- the
model itself never trains on it.

\begin{description}\itemsep2pt
\item[Maia-3 (frozen).] The external 23M-parameter Chessformer human-move
prior \citep{maia3chessformer}, used unmodified throughout; it consumes the position, the last eight
positions, and both players' ratings.
\item[No-engine variant (\texttt{sf\_smoke\_twinA}).] The re-ranker with
the engine feature block forced to zero, trained on the verified-human
$2800+$ rows of the $2500+$ pool and its remainder (rated Blitz and Rapid,
2025-06 to 2026-05). One side of the seed-paired ablation pair.
\item[$+$Stockfish variant (\texttt{sf\_smoke\_twinB}) and Lc0 variant
(\texttt{sf\_smoke\_twinB\_lc0}).] Identical to the no-engine variant in
data, seeds, and schedule, differing only in the engine block: Stockfish
scores (depth $14$/$10$s, \texttt{multipv}$=$$16$) or Lc0 scores \citep{lc0} (v0.31.2,
$800$ nodes). Their differences from the no-engine variant are the engine marginals of
the main paper.
\item[\textsc{Base} (\texttt{base\_hi}).] The general-mix intermediate:
trained from scratch (seed $0$, four epochs, rapid oversampled $4\times$)
on the $2500+$ pool support, the pool remainder, the GM-tranche
($2800+$) support, and an anonymous low-Elo regularizer -- all
verified-human rows, with Stockfish candidate features where annotated.
\item[\textsc{Matilda} (\texttt{base\_3k}).] The main paper's model:
\textsc{Base} warm-started for three epochs (learning rate $10^{-4}$) on
the verified-human rows of the dedicated mover-$\geq$$3000$ tranche (24
months) plus a deterministic $10\%$ replay of the general mix.
\item[Style vectors (\texttt{style\_token\_3k}).] The $32$-d per-player
embedding table and its zero-initialized projection, trained on the frozen
\textsc{Matilda} over the pool and GM-tranche support ($6{,}998$ distinct players;
embedding dropout $0.1$ trains the generic row). The \textsc{Base}-era
analogue (\texttt{style\_token\_hi}) appears only in the promotion
comparison below.
\item[Post-hoc club embeddings (\texttt{posthoc\_lostyle\_3k}, k-shot
variants).] New embedding rows for $1000$--$2500$ players fitted with the
base \emph{and} the style projection frozen, on the club-tranche support;
the k-shot variants fit on only the first $k$ decisions of fully held-out
players.
\item[Pre-coverage variants (\texttt{*\_pre3b}).] The \textsc{Base} recipe
trained before the GM-tranche support received engine annotations -- the
annotation-coverage ablation.
\end{description}

Every checkpoint above is released with a loader README and a manifest
recording its exact producer command and training data; the evaluation
rows for all comparisons are the main paper's verified-human benchmark
splits.

\section{Where the gains live: qualitative breakdowns}\label{app:quali}
Every result here is a metadata slice over the same per-row evaluations as the
main tables (no additional model runs; $95\%$ player-stratified bootstrap CIs
throughout; all on the human benchmark). Tags come from a single audited
tagging pass over (position, played move) with python-chess semantics
\citep{pythonchess};
sacrifice is deliberately untagged: unlike the tags above it is not
decidable from the (position, played move) pair alone. Whether
material offered is genuinely given up or won back by force depends on
search depth, and ``compensation'' is a judgment call, so any rule we
wrote could not be mechanically audited the way the other tags are.

\subsection{Game phase and ply} 
\begin{figure}[htbp]
\centering
\includegraphics[width=0.85\linewidth]{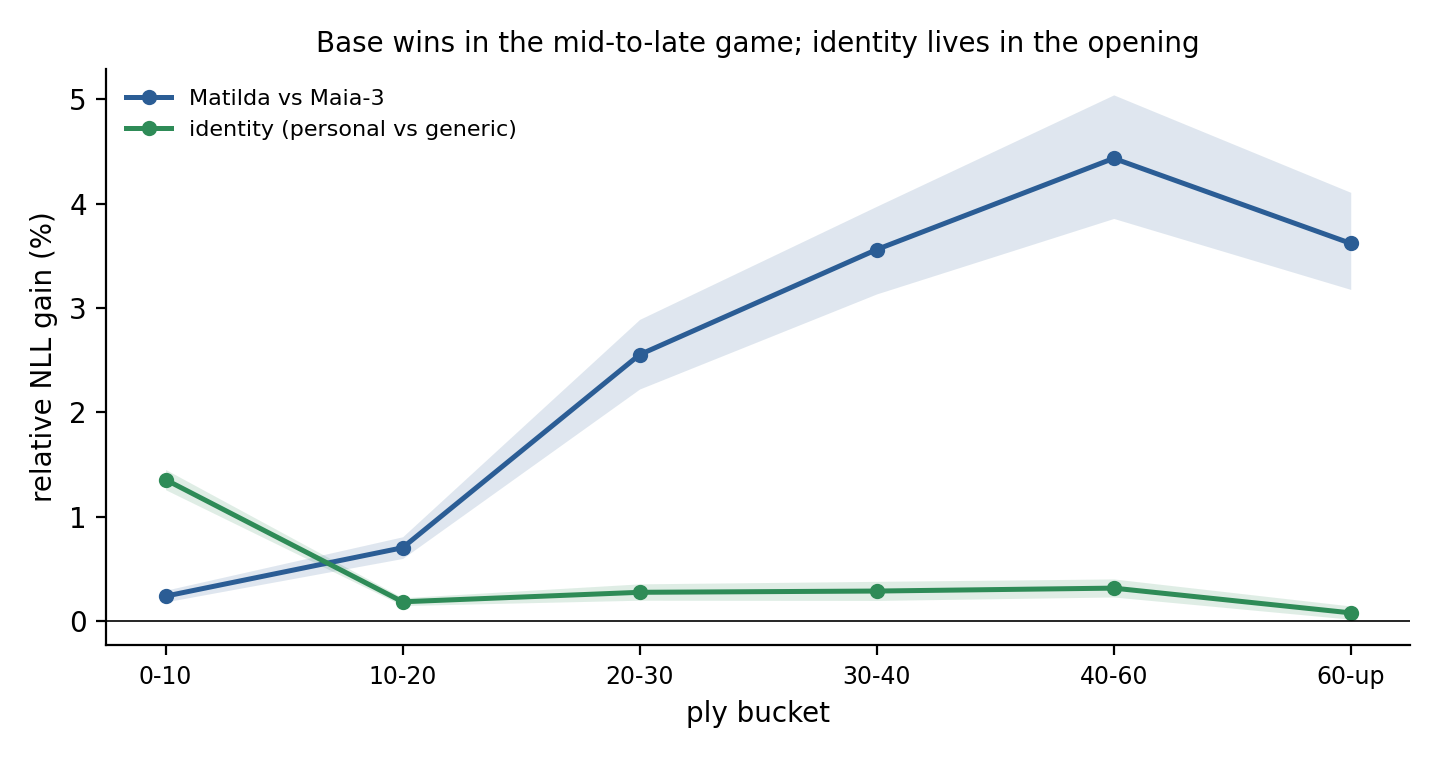}
\caption{Gain by ply bucket: \textsc{Matilda} (blue) vs identity (green), CI bands.}
\label{fig:plycurves}
\end{figure}

\textsc{Matilda}'s gains grow through the game: $+0.24\%$ in the
first ten plies (i.e.\ plies $2$--$9$; ply $<2$ is
excluded at extraction, the main paper's Data section) to $+3$--$4\%$ in the middlegame and
beyond -- while the identity gain concentrates in the opening
($+1.35\%$ at ply${<}10$ vs.\ $\sim$$+0.2\%$ after). See Figure~\ref{fig:plycurves}.

\subsection{Conditioning on the prior's rank} 
\begin{figure}[htbp]
\centering
\includegraphics[width=0.7\linewidth]{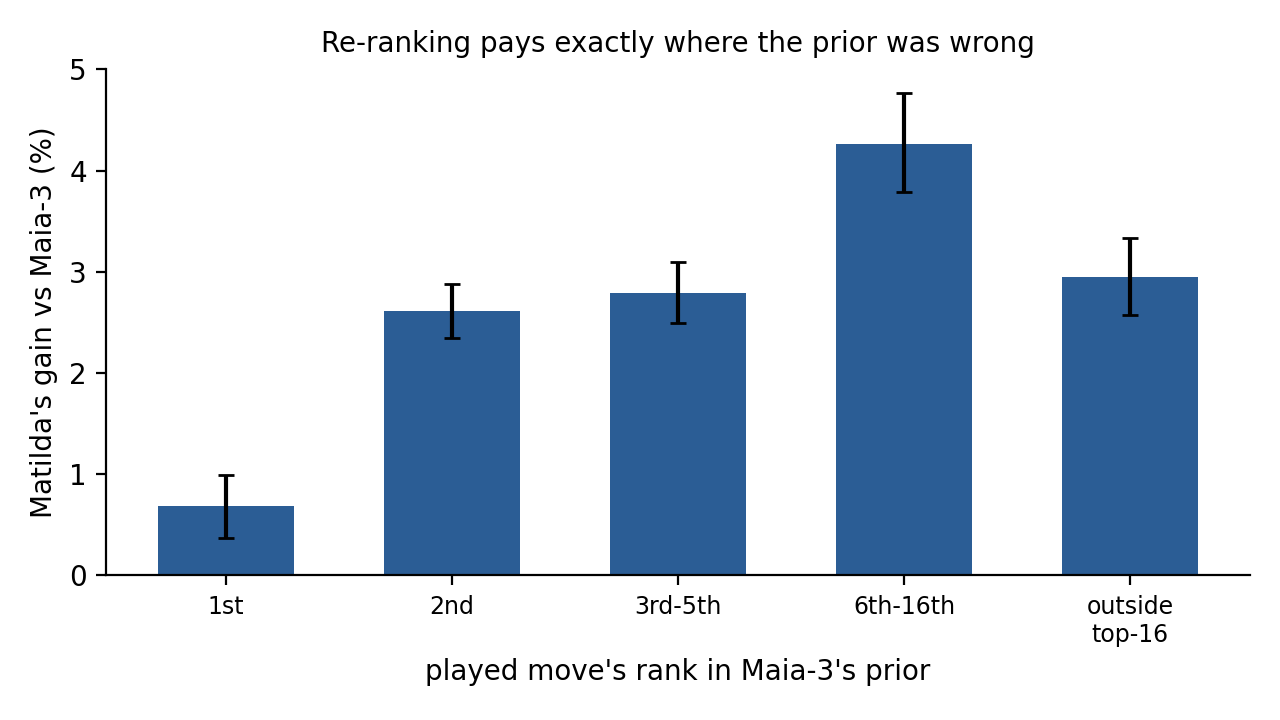}
\caption{Re-ranking pays where the prior was wrong.}
\label{fig:rankcond}
\end{figure}

Bucketing by the played move's
rank in Maia-3's prior \citep{maia3chessformer} (Table~\ref{tab:rankcond},
Figure~\ref{fig:rankcond}): when the prior was already right (rank 1,
$62\%$ of rows) re-ranking adds little ($+0.7\%$ -- the intermediate
general-mix checkpoint was slightly negative here, hedging probability
mass), and gains grow monotonically as the prior's rank worsens. Rows whose
played move fell outside the top-$16$ ($0.16\%$) still gain
$+3.0\%$: with no residual available on the played
move itself, this gain comes purely from demoting wrong candidates
under the full-vocabulary softmax.

\begin{table}[htbp]
\centering
\caption{\textsc{Matilda}'s gain conditioned on the played move's Maia-3 rank.}
\label{tab:rankcond}
\begin{tabular}{lrc}
\toprule
played move's Maia-3 rank & $n$ & base gain \\
\midrule
1st & 967,518 & $+0.69$ [+0.37,+0.99] \\
2nd & 284,706 & $+2.61$ [+2.35,+2.88] \\
3rd--5th & 233,901 & $+2.79$ [+2.50,+3.10] \\
6th--16th & 71,224 & $+4.27$ [+3.79,+4.77] \\
outside top-16 & 2,436 & $+2.95$ [+2.57,+3.33] \\
\bottomrule
\end{tabular}

\end{table}

\subsection{Move classes} Table~\ref{tab:movecls} (the companion
figure is in the main paper): \textsc{Matilda} is strongest on tactically pointed moves
(knight threats $+5.7\%$, checks $+4.1\%$, rook-to-open-file $+3.6\%$)
and essentially neutral on checkmates ($+0.2\%$, $n{=}2{,}027$; the
intermediate checkpoint was $-5.7\%$ -- the prior already handles most mates
and a weaker residual hedges). Identity is largest
on quiet pawn moves and castling (repertoire-adjacent choices), consistent
with the opening-heavy ply profile.

\begin{table}[htbp]
\centering
\caption{Gains by move class (multi-label; python-chess definitions in the
released tagger, self-checked).}
\label{tab:movecls}
\begin{tabular}{lrcc}
\toprule
move class & $n$ & base gain & identity gain \\
\midrule
quiet pawn move & 348,242 & $+1.64$ [+1.41,+1.88] & $+0.64$ [+0.58,+0.71] \\
castling & 43,674 & $+0.81$ [+0.57,+1.07] & $+0.76$ [+0.56,+0.97] \\
checking move & 89,192 & $+4.11$ [+3.66,+4.50] & $+0.18$ [+0.06,+0.30] \\
checkmate & 2,027 & $+0.18$ [-1.61,+2.04] & $-0.88$ [-1.76,-0.03] \\
rook to open file & 46,080 & $+3.60$ [+3.07,+4.17] & $+0.14$ [+0.00,+0.27] \\
rook to semi-open file & 32,814 & $+2.46$ [+1.98,+2.97] & $+0.21$ [+0.08,+0.35] \\
knight threatens R/Q & 23,700 & $+5.74$ [+5.06,+6.44] & $+0.38$ [+0.20,+0.56] \\
knight fork & 8,274 & $+5.34$ [+4.49,+6.23] & $+0.12$ [-0.09,+0.35] \\
knight defends attacked piece & 10,416 & $+3.19$ [+2.62,+3.79] & $+0.51$ [+0.34,+0.70] \\
\bottomrule
\end{tabular}

\end{table}

\subsection{New-player learning curve (k-shot)}
\begin{figure}[htbp]
\centering
\includegraphics[width=0.8\linewidth]{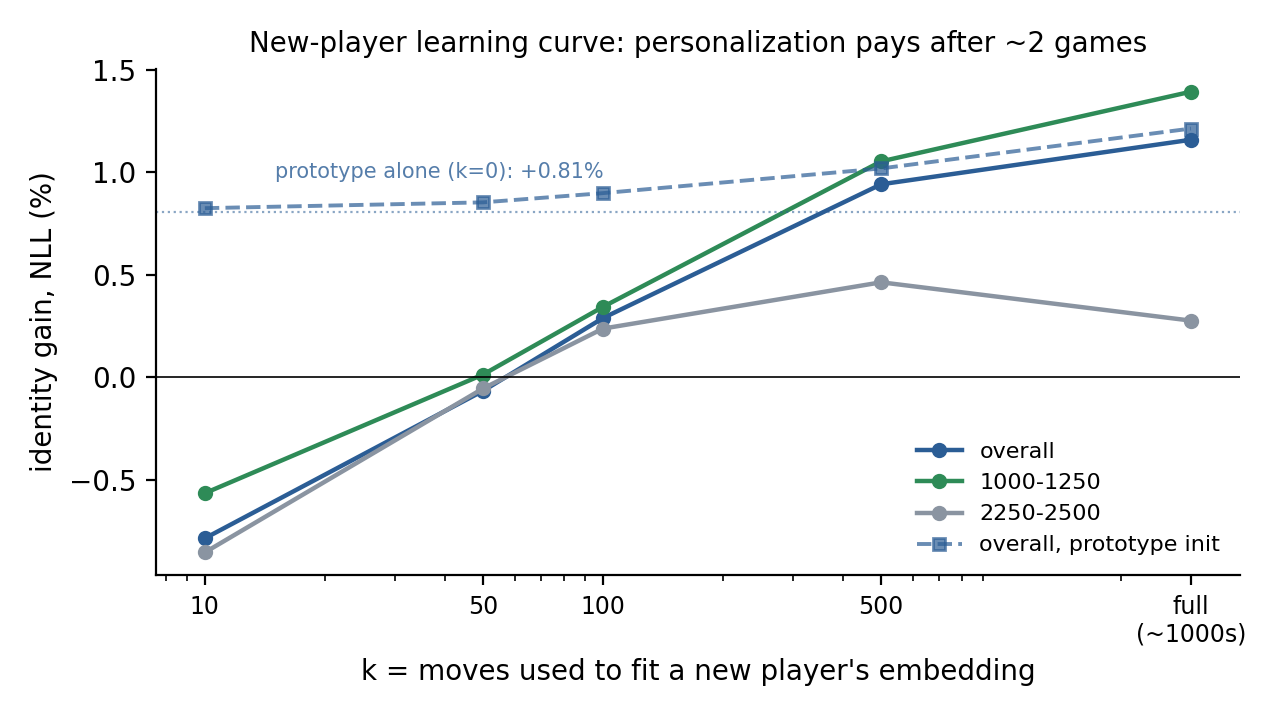}
\caption{k-shot new-player learning curve (fully held-out players;
embedding fit on the first $k$ moves, evaluated on the remainder).
Dashed: the same sweep under Maia4All-style prototype initialization which 
removes cold starts, which causes $4\times$ the rating leakage (see text).}
\label{fig:kshot}
\end{figure}

On players held out from all training, fitting an embedding on each
player's first $k$ decisions overfits at $k{=}10$ ($-0.79\%$), breaks
even near $k{=}60$, and reaches $+0.29\%$ at $k{=}100$ and $+0.94\%$
at $k{=}500$ (Figure~\ref{fig:kshot}): personalization helps after
roughly two games. Initializing new rows at Maia4All-style ability
prototypes -- $250$-Elo-band centroids of the fitted club embeddings,
with held-out players never contributing -- removes the cold start:
$+0.81\%$ with no fitting and $+0.83\%$ at $k{=}10$ (dashed). The
price is strength entanglement: the rating-recovery $R^2$ rises to $0.48$, four
times the zero-initialized $0.12$, perhaps because an ability prototype is
strength information by construction. Zero initialization keeps the
style space clean and costs a two-game warm-up.

\subsection{The mechanism}
\begin{figure}[htbp]
\centering
\includegraphics[width=0.62\linewidth]{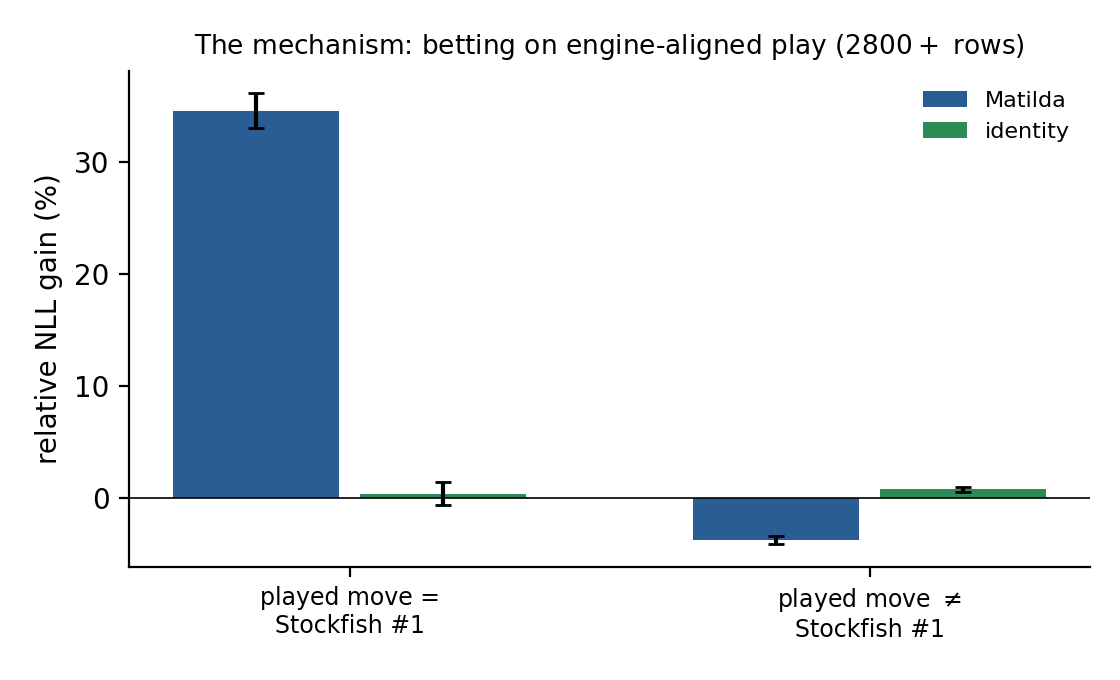}
\caption{Gain split by played-move agreement with Stockfish ($2800+$ rows).}
\label{fig:enginesplit}
\end{figure}

Table~\ref{tab:engdecomp} splits the seed-paired engine marginal on
engine-annotated ($2800+$) rows two ways at once: was the played move
Stockfish's top choice, and was the prior's own top choice
refuted -- scored at least $100$cp below the engine's best
candidate? The refuted slice is the sharpest evidence of expert
calculation: the marginal's largest payoff, $+41\%$, comes on rows
where the prior's favorite move fails tactically and the human found
the engine-approved alternative, and humans do avoid the prior's
refuted favorites ($69\%$ at $2800+$, $83\%$ at $3000+$, versus a
$35\%$ baseline) -- experts catch the tactical error the pattern prior
misses. The payoff concentrates on the engine's own choice: on rows
where the human avoided the refuted favorite but played some third
move, the marginal is negative ($-3.8\%$). That concentration is what
expert calculation should look like -- de Groot's protocols found that
even masters verify only a few candidate moves deeply
\citep{degroot1965thought}, and when that narrow search runs it lands
on the objectively best candidate -- so demoting the prior's mistake
and promoting the engine's choice are one and the same learned update.
Collapsed to agreement alone (Figure~\ref{fig:enginesplit}), the full
model gains $+34.5\%$ where the human played the engine move and is
$-3.8\%$ worse than Maia-3 where they deviated: not blind convergence
to the engine distribution, but tactical correction priced by how often
expert calculation reaches the engine's choice. Identity remains
positive on both sides (including $+0.74\%$ on non-engine moves: a
clean style effect, not engine-following). The same decomposition
reproduces on Go professionals (Go appendix).

\begin{table}[htbp]
\centering
\caption{The $2800+$ engine marginal by mechanism slice: (played move
$=$ engine top-1?) $\times$ (prior's top choice refuted at
$\geq$$100$cp?). Shares are of the total NLL reduction; the negative
slices make the positive shares exceed one. From
\texttt{eval/eval\_engine\_decomp.py}.}
\label{tab:engdecomp}
{\small\setlength{\tabcolsep}{3pt}
\begin{tabular}{lrrr}
\toprule
slice & $n$ & engine marginal \% & share of reduction \\
\midrule
played engine top-1, prior's choice refuted & 886 & +41.3 & +0.38 \\
played engine top-1, prior's choice sound & 23,317 & +21.4 & +1.50 \\
deviated, prior's choice refuted & 1,673 & -3.8 & -0.06 \\
deviated, prior's choice sound & 23,456 & -4.2 & -0.82 \\
\bottomrule
\end{tabular}
}
\end{table}

\subsection{Robustness of the aggregate} Per-player ($n{=}5{,}734$ with
$\geq$$100$ eval rows): $62.7\%$ of players improve under
\textsc{Matilda} relative to Maia-3, and $61.5\%$ improve further
under their personal embedding relative to the generic one. Per-player row counts are tightly capped in the
$2500$--$2800$ pool bands (top-3 players hold $\leq$$0.7\%$ of any band's
rows) but not in the GM/elite tranches (top-10 players hold $29.5\%$ of
$2900$--$3000$ rows and $39.1\%$ of $3000+$ rows); the per-player
interquartile means (IQMs)
reported alongside every row-weighted band estimate are the
concentration-robust view, and at $3000+$ they agree ($+20.5$ vs $+21.9$) -- indicating 
the gains are broadly shared across the players rather than just a few. Time
pressure helps \textsc{Matilda} ($+4.1\%$ under $10$s vs $+2.6\%$ over
$60$s) while identity needs time ($-0.3\%$ vs $+0.6\%$): style seems to disappear
in the scramble. Conditioning on the prior's entropy is a null result (
$+3.1\%$ most-confident quartile vs $+2.9\%$ least) -- gains do not
simply concentrate where the prior is diffuse.

\subsection{Famous-player case studies} 
\begin{figure}[htbp]
\centering
\includegraphics[width=0.9\linewidth]{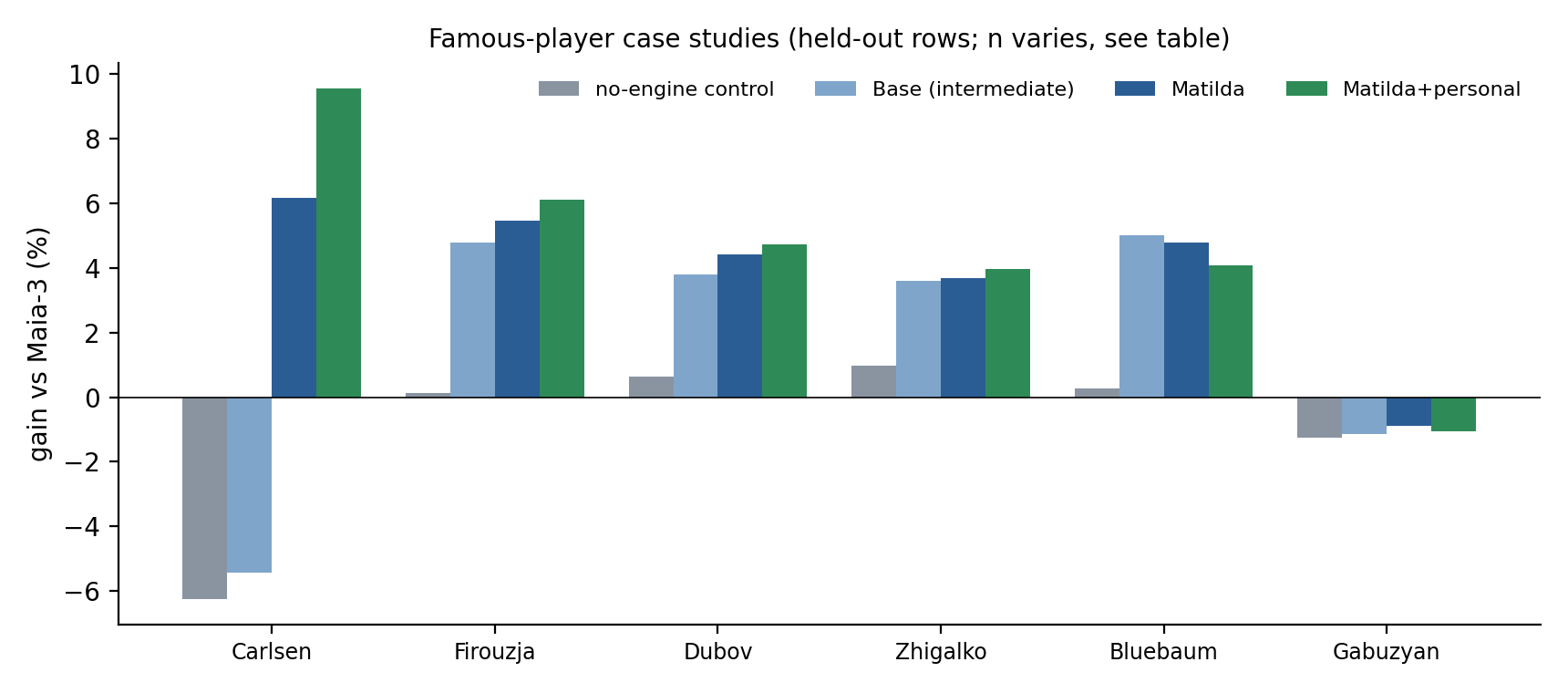}
\caption{Tier progression per famous player.}
\label{fig:famous}
\end{figure}

Table~\ref{tab:famous} and
Figure~\ref{fig:famous}: six GM accounts whose real-world attribution is
public (titled accounts under the player's own name or universally reported).
The standout is Magnus Carlsen: the intermediate general-mix checkpoint is
worse than Maia-3 on him ($-5.4\%$ -- the world champion is
out-of-distribution even for a GM-trained model), while \textsc{Matilda}
gains $+6.2\%$ and \textsc{Matilda}$+$personal reaches $+9.6\%$ -- the
clearest single illustration of why the final $3000+$ training stage
exists, and of personalization compounding on top of it. Thin slices are reported
as-is (GM Gabuzyan has only $n{=}108$ held-out rows).

\begin{table}[htbp]
\centering
\caption{Famous-player tier progression on each player's held-out rows
(relative NLL improvement vs Maia-3, \%, as in all main tables):
no-engine variant $\to$ intermediate checkpoint
(\textsc{Base}) $\to$ \textsc{Matilda} $\to$ \textsc{Matilda} with the
player's style vector. Attributions: titled accounts under the player's own
name/brand or universally reported (e.g.\ Titled Arena coverage). Rows are
within-player correlated; case studies, not population estimates.}
\label{tab:famous}
\begin{tabular}{llrrcccc}
\toprule
player & account & $n$ & Elo & no-eng & \textsc{Base} & \textsc{Matilda} & $+$style \\
\midrule
Magnus Carlsen & \texttt{DrNykterstein} & 373 & 3121 & $-6.2$ & $-5.4$ & $+6.2$ & $+9.6$ \\
Alireza Firouzja & \texttt{alireza2003} & 211 & 2918 & $+0.1$ & $+4.8$ & $+5.5$ & $+6.1$ \\
Daniil Dubov & \texttt{Vladimirovich9000} & 1,997 & 2916 & $+0.6$ & $+3.8$ & $+4.4$ & $+4.7$ \\
Sergei Zhigalko & \texttt{Zhigalko\_Sergei} & 2,311 & 2824 & $+1.0$ & $+3.6$ & $+3.7$ & $+4.0$ \\
Matthias Bluebaum & \texttt{Msb2} & 396 & 2833 & $+0.3$ & $+5.0$ & $+4.8$ & $+4.1$ \\
Hovhannes Gabuzyan & \texttt{GABUZYAN\_CHESSMOOD} & 108 & 2697 & $-1.3$ & $-1.1$ & $-0.9$ & $-1.1$ \\
\bottomrule
\end{tabular}

\end{table}

\begin{figure}[htbp]
\centering
\includegraphics[width=0.9\linewidth]{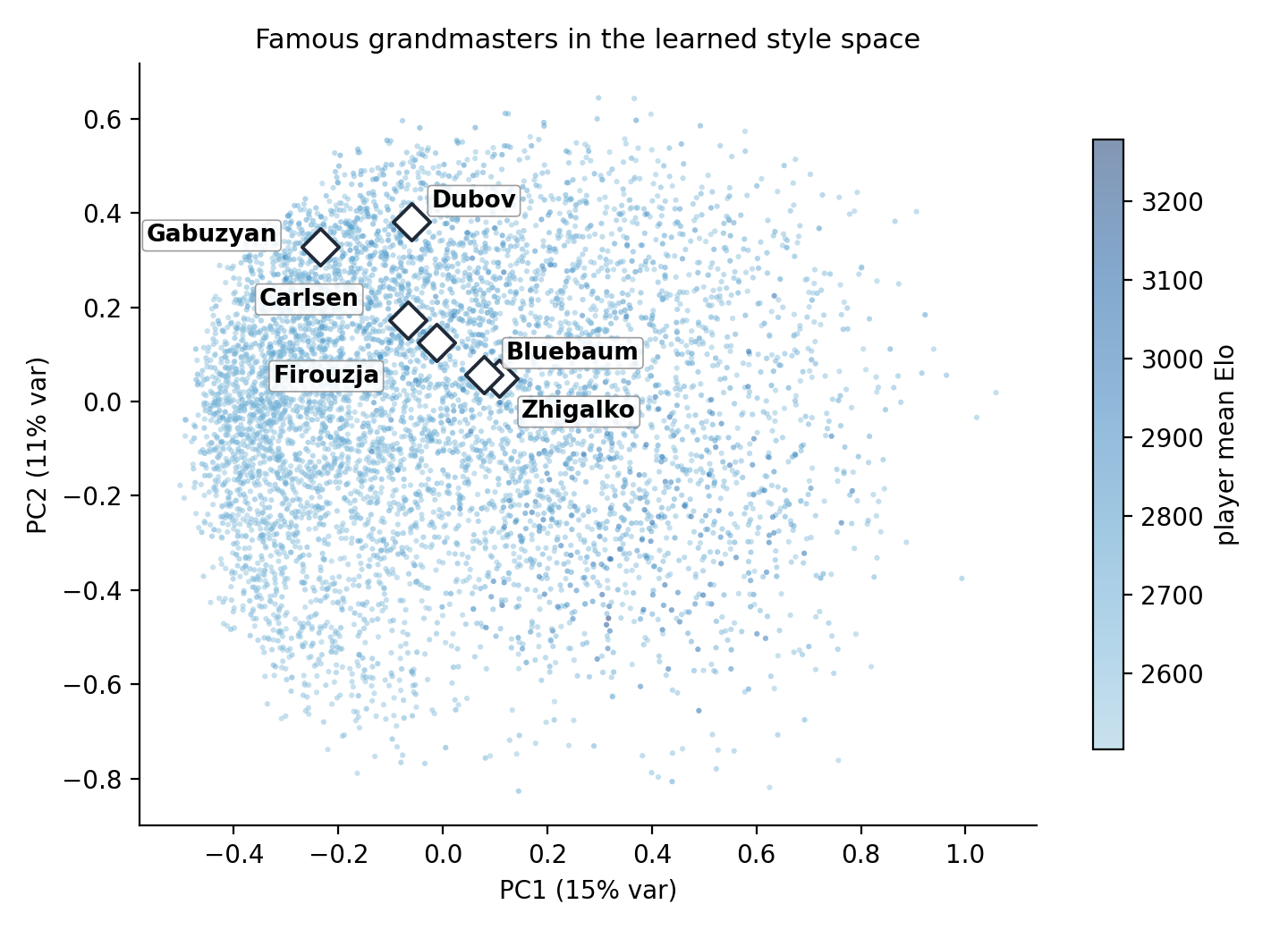}
\caption{Six publicly-attributed grandmaster accounts inside the learned
style space (PCA of all trained $2500+$ style vectors; background colored
by player mean Elo, as in the main paper's Elo-disentanglement figure). The accounts 
land in a coherent region of the space yet
spread within it.}
\label{fig:famousspace}
\end{figure}

Figure~\ref{fig:famousspace} additionally places the six accounts inside
the learned style space: they land in a coherent region of the space yet
spread within it. This is consistent with the vectors encoding shared
strong-player structure, especially for elite play where the elo conditioning 
that must flow through Maia in the architecture runs out, plus individual identity.

\FloatBarrier

\subsection{Absolute performance across the ladder}
Table~\ref{tab:e8} reports absolute numbers across the tested ladder
($1000$--$4000$). Our baseline is consistent with Maia-3's published
move-matching ($56.6\%$ for the 23M model we build on, aggregated over
their full ladder; on our $1000$--$2500$ evaluation its top-$1$ is
$57.1\%$, and stronger populations are more predictable). At $3000+$
the model moves top-$1$ accuracy from $61.1\%$ to $68.4\%$ while
cutting NLL by nearly a quarter. In the $2500+$ aggregate the style
vector moves NLL but barely moves top-$1$: style is a preference among
reasonable moves, not a different notion of the best one.

\begin{table*}[htbp]
\centering
\begin{tabular}{lcccccccc}
\toprule
 & \multicolumn{2}{c}{$1000+$ (all tested)} & \multicolumn{2}{c}{$2500+$} & \multicolumn{2}{c}{$2800+$} & \multicolumn{2}{c}{$3000+$} \\
method & NLL & top-1 & NLL & top-1 & NLL & top-1 & NLL & top-1 \\
\midrule
Maia-3 & 1.139 & 0.610 & 1.098 & 0.620 & 1.150 & 0.606 & 1.165 & 0.611 \\
\textsc{Matilda} & 1.117 & 0.615 & 1.070 & 0.628 & 1.066 & 0.630 & 0.910 & 0.682 \\
\textsc{Matilda}$+$style & 1.114 & 0.616 & 1.066 & 0.628 & 1.058 & 0.632 & 0.899 & 0.684 \\
\bottomrule
\end{tabular}

\caption{Absolute performance on held-out decisions (full legal-move
scoring): NLL (lower is better) and top-$1$ accuracy (higher is better) per
rating range. The $1000+$ column spans the whole tested ladder (the
anonymous $1000$--$2500$ evaluation plus the $2500+$ benchmark); its rows
have no player identities, so the style row scores under the generic
fallback there.}
\label{tab:e8}
\end{table*}

\section{Verification, audits, and reproduction}\label{app:audits}
\subsection{Baseline validation} Maia-3 (23M) is the anchor; the choice over
Maia-2 and over the 79M variant, and a check of our Maia-3 integration against the accuracies
published on the ALLIE test set \citep{zhang2025allie}, are summarized in
Table~\ref{tab:maia2-vs-maia3}.
\begin{table}[htbp]
\centering
\caption{Top-1 move-matching accuracy by rating: Maia-2 rapid vs.\ Maia-3
(23M/79M), same held-out decisions with reconstructed 8-ply history. Our Maia-3
numbers bracket the accuracies \citet{maia3chessformer} report on the {ALLIE}
\citep{zhang2025allie} blitz test set ($56.6\%$ for 23M, $57.1\%$ for 79M; their protocol removes the
first 10 plies and sub-30s-clock moves, ours does not), validating our Maia-3 integration.
Note Maia-3 is trained on blitz games only and Maia-2's released model
is rapid-trained, while these decisions are rapid -- both baselines are
therefore partially out-of-domain here, in opposite directions.}
\label{tab:maia2-vs-maia3}
\begin{tabular}{lccc}
\toprule
rating bucket & Maia-2 & Maia3-23M & Maia3-79M \\
\midrule
900--1200  & 0.487 & 0.519 & 0.521 \\
1500--1800 & 0.513 & 0.546 & 0.551 \\
2100--2300 & 0.518 & 0.568 & 0.588 \\
2500--2800 & 0.520 & 0.588 & 0.612 \\
2800+      & 0.485 & 0.568 & \textbf{0.580} \\
\bottomrule
\end{tabular}
\end{table}

Every headline number is re-verified empirically before reporting; the
checks below are automated and run as part of the reproduction pipeline.

\subsection{Tranche selection details} The main paper's tranche table
compresses the selection rules; this is the full set. Named-player tranches keep
only players with enough history ($\geq600$ decisions for the pool,
$\geq100$ for GM and $3000+$, $\geq300$ for club, whose players are first
hash-sampled $1$-in-$500$). Per-player caps then bound contributions: a
$1{,}500$-row total across the window for the pool, keeping it broad;
because elite play is scarce, the GM and $3000+$ tranches instead cap per
month ($3{,}000$ and $6{,}000$ rows/player/month), letting prolific
strong players contribute more. Club players are uncapped. Player counts:
$6{,}237{+}692$ (pool, train $+$ held out), $1{,}458{+}87$ (GM), $532$
raw accounts at $3000+$ (mostly engines/cheaters before the audit; $378$
qualify, $154$ verified-human), $4{,}038{+}462$ (club, $1000$--$2500$;
the $462$ held out are the k-shot new-player set); the anonymous low-Elo
sets span $12{,}097$ (eval) and $39{,}763$ (train) distinct movers,
scored without identities. All counts are the raw extraction: the
account-title audit's bot/TOS exclusion is applied in the data loaders,
before all training and evaluation.

\subsection{Account-title audit} Every account we train
or evaluate on was resolved against the Lichess users API: $1{,}127$ carry the
BOT title (engine accounts) and $1{,}777$ are TOS-flagged (marked cheaters).
They dominate the public high-rating population -- $59$--$77\%$ of an
unfiltered $2900$--$3000$ extraction, $76$--$98\%$ of the raw $3000+$
tranches -- while the $1000$--$2500$ club tranche is $0.0\%$ BOT. All are
excluded via name-based per-file exclusion masks applied identically at training and
evaluation; a conformance suite verifies the masks functionally, and every
evaluation artifact carries a provenance stamp of the exclusion list it was
computed under, so non-conforming numbers cannot silently enter the
canonical tables.

\subsection{No label injection} Candidates are Maia-3's own top-$16$; the
played move is never added. Measured target-in-candidate coverage is
$99.87\%$ at $2500$--$2800$ declining to $99.69\%$ at $3000+$ --
genuine misses that pattern with Maia-3's out-of-domain error, where an
injected set would read exactly $100\%$.

\subsection{Full-vocabulary normalization, numerically} On real evaluation
rows the model's probability mass sums to $1.000000$ over the full $4352$-move
vocabulary, illegal moves carry exactly zero mass, and rows whose played move
falls outside the candidates still receive non-zero probability through the
Maia-3 prior (mean $1.75\%$ on such rows) -- a residual on the full
distribution, not a candidate softmax.

\subsection{Seen-position audit (unseen-key slice)} Temporal per-player splits
let a player's recurring openings -- and at $3000+$, recurring deep theory
between the same few players -- place an exact
$(\text{player},\text{position},\text{move})$ key on both sides of the split:
$10.3\%$ of style-pool eval rows ($34.6\%$ at ply${<}10$, $0.7\%$ past ply
$20$), with deep-theory repeats concentrated at the top. Excluding
every such row: $+4.3\%$ / $+11.6\%$ / $+20.9\%$ at $2800$--$2900$ /
$2900$--$3000$ / $3000+$ (vs.\ $+4.3/+11.9/+21.9$ on all human rows).
Repertoire memorization is not the driver; arguably the seen slice is itself
style signal, but the unseen-key numbers stand on their own.

\subsection{Game-overlap audit (headline survives)} A subtler channel:
when two pool players meet, one side's moves can train while the other side's
moves from the same game are evaluated -- disjoint positions and
targets, but the training rows' history features encode the eval player's
replies. A game-identity re-extraction (deterministic, verified row-identical
to the original tranches) quantifies it: games where only the
opponent's side trained cover just a few percent of eval rows. The maximally
conservative slice -- excluding every eval row whose
$(\text{player},\text{position},\text{move})$ key appears in any
game that contributed any training row, either side, which subsumes
both the repeated-position channel and games straddling the train/eval
boundary -- still gives $+3.8\%$ at
$2800$--$2900$, $+10.7\%$ at $2900$--$3000$, and $+24.0\%$ [$17.8,30.8$]
at $3000+$ -- statistically indistinguishable from the all-rows gains
(the $3000+$ slice retains $3{,}510$ rows over $43$ players). The
seed-paired $\pm$engine variants (the main paper's Stockfish-ablation
section) are immune by construction and
corroborate the attribution.

\subsection{Matched-protocol slice (gains survive)} Maia-3's published
evaluation protocol \citep{maia3chessformer} drops the first ten
moves of each game (twenty plies) and every position after either
player first falls under $30$ seconds. Our slice approximates it as ply
$\geq 10$ and per-position mover clock $\geq 30$s -- a strictly weaker
filter in both respects, so it under- rather than over-corrects. Our main
tables keep these rows (they carry style and time-pressure signal). Re-evaluating under
this reference convention: $+5.0\%$ at $2800$--$2900$, $+14.4\%$ at
$2900$--$3000$, $+24.2\%$ [$19.8,28.5$] at $3000+$. These point estimates sit
directionally above the all-rows gains (opening theory is where Maia-3 is
relatively strongest) but within the player-bootstrap intervals. The
mixed-convention numbers in the main text remain the conservative choice.

\subsection{Rating-input handling (no clipping above our range)} Because
the top bands are where our gains concentrate, we verify Maia-3 receives
high ratings faithfully. Its rating pathway clamps inputs to $[0,5000]$ --
far above our $3300$ maximum -- and embeds them by continuous linear
interpolation between two learned endpoint embeddings; there is no
inference-time binning, and our pipeline feeds exact integer ratings, so
$3000+$ rows are conditioned on their true ratings. A deterministic probe
(\texttt{eval/probe\_elo\_response.py}: three fixed positions, both
ratings swept $2400$--$3200$) shows the learned response does continue to
move above $2600$ -- total-variation distance from the $2600$ output grows
to $0.05$--$0.10$ by $3200$ in the opening and endgame positions -- but the
movement can be poorly grounded: in the middlegame position the top-move
probability collapses from $0.65$ to $0.27$ at $3200$ (TV $0.39$), a swing
the training signal cannot support: Maia-3 balances its training data
over $100$-point rating bins up to $2600$, and all higher-rated games
share one final bin under the same per-bin quota -- the entire $2600+$
range contributes no more data than a single $100$-point band, and
nothing in that bin distinguishes $2700$ from $3200$. This motivates the
main text's wording that the rating
conditioning is unreliable above $2600$: the input is not
clipped, but the learned mapping there is thin -- exactly the regime the
engine features repair.

\subsection{Engine-signal robustness} Two checks that the engine features
are consumed generically rather than tuned to one engine's idiosyncrasies.
(i) Depth. A $181$k-row sample was annotated at both the production
setting (depth $14$/$10$s) and depth $21$/$30$s; all statistics below are on
its $89{,}692$ verified-human rows (the raw dual re-search targets $2800+$,
where BOT/TOS accounts are approximately half the population). The rankings
agree strongly (rank correlation $0.869$; same best candidate $67.8\%$ of
rows), and substituting the deeper features zero-shot into the d14-trained
\textsc{Base} leaves the gain essentially unchanged at $2800$--$3000$
($+0.02$pp) with $+1.9$pp of headroom at $3000+$
(Table~\ref{tab:dualdepth}, $n{=}6{,}426$ there) -- the model reads engine
signal monotonically, the cheap annotation was enough below $3000$, and the
production numbers are, if anything, conservative in engine strength at the
very top.
(ii) Engine family. A third seed-paired variant trained on Lc0
annotations (MCTS $+$ policy network, a different paradigm from Stockfish's
alpha-beta $+$ NNUE; fixed $800$ nodes, pinned net) tests whether the
Dimension-1 gradient depends on how the engine thinks
(Table~\ref{tab:engines}).

\begin{table}[htbp]
\centering
\caption{Depth robustness on the dual-annotated sample: \textsc{Base}
(trained on d14 features) evaluated with d14 vs.\ zero-shot d21 features,
identical rows.}
\label{tab:dualdepth}
\begin{tabular}{lrccc}
\toprule
Elo band & $n$ & d14 features (as trained) & d21 substituted & $\Delta$ (pp) \\
\midrule
2800--3000 & 83,266 & $+4.7\%$ & $+4.8\%$ & $+0.0$ \\
3000--4000 & 6,426 & $+18.0\%$ & $+19.9\%$ & $+1.9$ \\
2800--4000 & 89,692 & $+5.8\%$ & $+5.9\%$ & $+0.2$ \\
\bottomrule
\end{tabular}

\end{table}

\ifstandalonesupplement
\begin{table}[htbp]
\centering
\caption{Engine-family robustness: seed-paired ablation marginals over
the no-engine variant, Stockfish vs.\ Lc0 features on identical data and
seeds.}
\label{tab:engines}
\begin{tabular*}{\columnwidth}{@{\extracolsep{\fill}}lrcc@{}}
\toprule
Elo band & $n$ & Stockfish marginal & Lc0 marginal \\
\midrule
2800--2900 & 35,528 & $+2.3\%$ & $+2.5\%$ \\
2900--3000 & 10,018 & $+5.1\%$ & $+5.1\%$ \\
3000--4000 & 3,881 & $+9.6\%$ & $+6.6\%$ \\
2800--4000 & 49,427 & $+3.4\%$ & $+3.3\%$ \\
\bottomrule
\end{tabular*}

\end{table}
\fi

\subsection{Search-allocation probe (the interface is learnable)} 
\begin{figure}[htbp]
\centering
\includegraphics[width=0.9\linewidth]{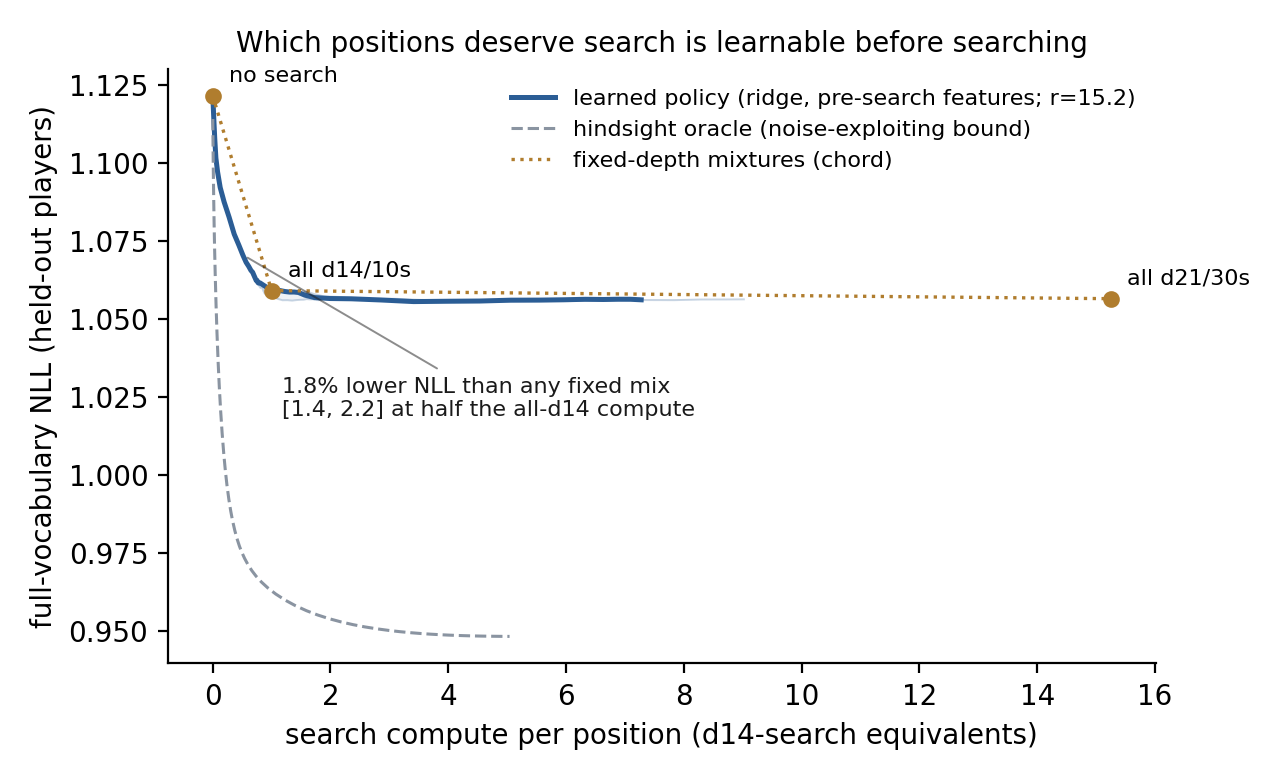}
\caption{Search-allocation probe: NLL vs.\ search compute on held-out
players. The learned pre-search allocation (blue) vs.\ fixed-depth mixtures
(the chord, red) and a hindsight oracle (gray); shaded band $=$ cost-ratio
sensitivity ($3$--$20\times$; measured $15.25\times$).}
\label{fig:searchpolicy}
\end{figure}

The
candidate-token contract turns per-position search allocation into an action
space; as a deliberately minimal feasibility probe, not a search
controller, we ask whether where to spend search is predictable
before searching. On the dual-depth sample's verified-human,
evaluation-side rows ($56{,}531$ decisions from $931$ players, none used to
train the base), we score \textsc{Matilda} under three evidence conditions
(no search, depth-$14$, depth-$21$), fit closed-form ridge regressors
predicting each option's NLL gain from strictly pre-search features (prior
shape, rating, clock, phase), and trace the compute--NLL frontier on
held-out players: each position takes the option $k$ maximizing
$\hat g_k-\lambda c_k$, where $\hat g_k$ is that option's predicted gain,
$c_k$ its compute cost in depth-$14$ units ($0$, $1$, and the measured
ratio below), and $\lambda$ -- a price per unit of compute -- is swept
from high to low to trace the curve.
Search gains are heterogeneous enough that the ranking pays: at half the
all-depth-$14$ compute the learned allocation captures $82\%$ of the
full-search gain where unranked mixing gets its cost share ($51\%$) --
$1.8\%$ lower NLL than the mixture chord [$1.4,2.2$], player-stratified
bootstrap -- and a hindsight oracle -- the same pricing rule fed each
position's realized gains instead of predictions -- shows substantial
further headroom (Figure~\ref{fig:searchpolicy}). The oracle is an
upper bound for any predictor, and a loose one: it also collects
positions where deeper search happened to win on these rows, which no
pre-search policy could know. The measured cost ratio between the two
depths is $15.25\times$ [$14.42,16.24$] (timed re-search of $300$ of the sample's
positions under the production worker configuration; the nominal time caps
would suggest $3\times$). The headline claims are insensitive to this
measurement: at half the depth-$14$ budget the policy chooses almost
entirely between no-search (cost $0$) and depth-$14$ (cost $1$), neither
of which involves the ratio; the claims that do involve the depth-$21$
option hold at every ratio in a $3$--$20\times$ band (the figure's shaded
region). Mean NLL also understates the stakes: search's corrections
concentrate in tactically pointed positions (the move-class breakdown,
Table~\ref{tab:movecls}), and at high Elo a single unlikely tactical
error can decide a game -- being right in exactly those positions is
worth more than an average-NLL accounting shows. We therefore see this
as a promising direction for future work: making the search component
not only more human-like but also more computationally efficient. In
the limit the two may be the same objective -- spending compute where a
human would spend attention.

\subsection{Move-tag conformance} The tagging pass behind
the qualitative-breakdown appendix hard-asserts its own consistency (checkmate
$\subseteq$ check; quiet-pawn $\cap$ check $=\emptyset$; castling moves a
king; the rank-1 share must match Maia-3's independently-measured top-1
accuracy) and records class base rates (castling $2.8\%$ of moves, mates
$0.13\%$) in a provenance file checked by the conformance suite.

\subsection{Style-embedding PCA spectrum} The main paper's disentanglement
figures plot the first two principal components of each population's $32$-d
style embeddings. Those two components carry only a minority of the total
variance: $31.1\%$ for the club post-hoc embeddings and $26.7\%$ for the
jointly-trained $2500+$ token -- so most of the stylistic signal lies beyond
the plotted plane; the spectrum has a long tail with no dominant axis
(Table~\ref{tab:pcaspectrum}). This does not affect the disentanglement
result: the rating-recovery probe ($R^2$) is fit on the complete $32$-d
embedding, not on this two-dimensional projection. It does mean the scatter
plots should be read as a low-variance view of a genuinely
high-dimensional space.

\begin{table}[htbp]
\centering
\caption{Style-embedding PCA variance spectrum: \% of total variance per
principal component (same populations, row selection, and per-row
$L^2$-normalization as the main paper's disentanglement figures). The two
components the figures plot capture a minority of the variance; the spectrum
decays slowly across the top $10$.}
\label{tab:pcaspectrum}
\begin{tabular}{lcc}
\toprule
principal component & club (post-hoc) & $2500+$ (token) \\
\midrule
PC1 & 21.8 & 14.5 \\
PC2 & 9.3 & 12.2 \\
PC3 & 6.3 & 9.0 \\
PC4 & 6.0 & 6.3 \\
PC5 & 5.1 & 6.0 \\
PC6 & 4.9 & 5.2 \\
PC7 & 4.4 & 4.7 \\
PC8 & 3.8 & 4.3 \\
PC9 & 3.6 & 3.7 \\
PC10 & 3.1 & 3.3 \\
\midrule
PC1--2 (plotted in figures) & 31.1 & 26.7 \\
PC1--10 (cumulative) & 68.4 & 69.4 \\
\bottomrule
\end{tabular}

\end{table}

\subsection{Reproduction} The code appendix's top-level
\texttt{REPRODUCE.md} is the chess entry point (four parts, from
checking the released checkpoints to full regeneration);
\texttt{go\_pipeline/REPRODUCE.md} covers the Go experiments the same
way. All data selection is seed-deterministic (a
verification script checks on-disk row counts against the documented lineage
exactly); a single idempotent command regenerates every table with its CIs, and
each released table names the command and source artifact that rebuilds it.
Confidence intervals are $95\%$ player-stratified bootstrap (the player set
resampled with replacement $2000$ times), following the methodology of
\citet{agarwal2021deep}. Every reported model is exactly one training run with the fixed
seed listed in the seeds table (the seed-paired $\pm$engine ablation is a
seed-paired pair, so its difference controls for the shared seed);
the bootstrap CIs quantify evaluation-set sampling variability, not
training-seed variability, which is bounded only by the reproducibility note
below and the paired design. Training on Apple-Silicon MPS reproduces to
$\sim$3 decimals; data and evaluation reproduce exactly.
Paper names map to released artifacts as: \textsc{Base} $=$
\texttt{base\_hi.pt}, \textsc{Base-noSF} $=$ \texttt{base\_h8\_tc.pt},
\textsc{Matilda} $=$ \texttt{base\_3k.pt}, its style vectors $=$
\texttt{style\_token\_3k.pt}; tranches: pool $=$ \texttt{style\_*}, pool
remainder $=$ \texttt{base\_extra}, GM $=$ \texttt{hi2800\_*}, $3000+$ $=$
\texttt{hi3k\_*}, club $=$ \texttt{lostyle\_*}, anonymous low-Elo $=$
\texttt{loelo\_*}.

\section{Training Methodology and Intermediate Checkpoints}
\label{app:training}
The main text reports one model -- \textsc{Matilda} (code
\texttt{base\_3k}), the final checkpoint of a staged construction plus
its style vectors. This appendix documents how it was produced: the staged
construction, the intermediate checkpoints it passed through, and the
evidence that each stage was sound. The intermediate artifacts are released
alongside the final ones.

\subsection{Why training proceeds in stages}
Two constraints shape the pipeline. First, data density collapses with
rating: the $2500+$ pool holds millions of decisions, but verified-human
$3000+$ play is scarce ($154$ accounts over 24 months once engine and
flagged accounts are removed). Second, the expensive step is engine
annotation, not training, so each stage gates the next expenditure: a
12-configuration sweep at small scale selected the compact architecture; the
seed-paired $\pm$Stockfish runs on the $2800+$ subset were the go/no-go
gate for the full annotation program (main text); only then was
\textsc{Base} trained from scratch on the fully annotated mix, and
\textsc{Matilda} warm-started from it. Every stage shares the same
architecture and seed discipline (all trainers seed $0$; identical shard
shuffles), so successive checkpoints are comparable.

\subsection{Checkpoint lineage}
\textsc{Base-noSF} (code \texttt{base\_h8\_tc}) is the TC-conditioned,
engine-free parent. The gate pair warm-starts from it by widening the
candidate projection with zero-initialized columns for the engine block --
a checkpoint-exact start, so variant~A equals the parent and variant~B differs
only by the engine features. \textsc{Base} (code \texttt{base\_hi}) is then
trained from scratch on the full mix (main pool $+$ pool remainder $+$ GM
support $+$ a low-Elo regularizer), with rapid rows oversampled
$4\times$ -- rapid is $\approx$$6\%$ of the pool yet outside blitz-trained
Maia-3's domain, and $4\times$ (to roughly a fifth of the training
stream) proved just enough signal to learn the regime without letting a
slice the prior has never seen dominate training.
\textsc{Matilda} (code \texttt{base\_3k}; staged name \textsc{Elite})
warm-starts from \textsc{Base} at
learning rate $10^{-4}$ for three epochs on the human rows of the dedicated
mover-$\geq$$3000$ tranche plus a deterministic $10\%$ replay of the general
mix. The style token trains on the frozen \textsc{Base}; new players
receive embeddings post-hoc with only their embedding rows trainable.

\subsection{Annotation-coverage ablation (wave 3b)}
\textsc{Base} was first trained with its $2800+$ GM-tranche support
rows unannotated (engine features present only on the main pool).
Completing the annotations ($4.5$M rows, ``wave 3b'') and retraining the
identical recipe moved the GM bands from $+3.4/+10.7/+19.9\%$ to
$+4.0/+11.3/+20.2\%$ and doubled the identity gain at $2800$--$2900$
($+0.16\to+0.35\%$). The pre-coverage checkpoints are preserved
(\texttt{*\_pre3b}) and the ablation re-evaluates on the current benchmark
with one command: engine features help most when the model also
trained with them on the population being evaluated.

\subsection{The final stage dominates everywhere}
The promotion of the final-stage checkpoint (\textsc{Matilda}, code
\texttt{base\_3k}) over the general-mix intermediate (\textsc{Base}, code
\texttt{base\_hi}) was gated on a direct comparison over the whole $2500+$
benchmark, including a style-vector retrain on the frozen final checkpoint.
Table~\ref{tab:basevsmatilda} shows the outcome: the final stage improves
every band and every column -- base gain, personalized gain,
and identity -- including the guard bands far below its training focus, and
its embeddings are better Elo-disentangled (token probe
$R^2$ $0.19\to0.16$, club post-hoc $0.14\to0.12$): the specialist stage
absorbs within-band strength into the base, leaving the style space cleaner.
Specialization cost nothing; the main text therefore reports one model.

\begin{table}[htbp]
\centering
{\small
\begin{tabular}{lcccc}
\toprule
 & \multicolumn{2}{c}{base gain vs Maia-3} & \multicolumn{2}{c}{identity} \\
Elo band & \textsc{Base} & \textsc{Matilda} & \textsc{Base} & \textsc{Matilda} \\
\midrule
2500--2600 & $+0.33$ & $+0.46$ & $+0.21$ & $+0.29$ \\
2600--2700 & $+0.49$ & $+0.58$ & $+0.22$ & $+0.29$ \\
2700--2800 & $+0.51$ & $+0.56$ & $+0.21$ & $+0.25$ \\
2800--2900 & $+4.01$ & $+4.26$ & $+0.35$ & $+0.47$ \\
2900--3000 & $+11.32$ & $+11.95$ & $+1.21$ & $+1.46$ \\
3000--4000 & $+20.19$ & $+21.91$ & $+1.01$ & $+1.12$ \\
2500+ overall & $+2.31$ & $+2.53$ & $+0.31$ & $+0.41$ \\
\bottomrule
\end{tabular}
}
\caption{Intermediate vs final checkpoint on the full human benchmark:
relative NLL gain vs Maia-3 and identity gain, per band. The final stage
(\textsc{Matilda}) dominates every cell.}
\label{tab:basevsmatilda}
\end{table}

\subsection{Warm-start guards (no forgetting)}
Specialization did not cost generality. While training the final stage we
re-scored, after every epoch, a fixed subset of the $2500+$ eval split --
watching for regression on the broad bands that the every-tenth-shard
replay is meant to protect. Nothing regressed, and end to end the guarded
band improved ($+0.33\%\to+0.46\%$ at $2500$--$2600$, \textsc{Base} to
\textsc{Matilda}) while the target band climbed from $+17.7\%$ to
$+23.0\%$ on the $3000+$ tranche's own eval split ($75$k rows over the
$154$ verified-human accounts -- a different, larger population than the
benchmark's $3000+$ band, hence the difference from the main table's
$+21.9\%$). Below the benchmark range, the same low-Elo preservation
evaluation used for \textsc{Base} (anonymous $432$k rows, $1000$--$2500$)
holds for the final checkpoint as well -- the basis for deploying it as a
single general-purpose model.

\subsection{Computing infrastructure}
Engine annotation runs on Google Cloud Run: candidate tasks are sharded
into parquet files on Cloud Storage and consumed by fleets of
single-vCPU containers -- up to $195$ concurrent Stockfish workers
(depth $14$/$10$s per position; $10.2$M positions searched across the
annotation waves) and $100$-way-parallel Lc0 workers ($299$ tasks,
$\approx$$1{,}730$ vCPU-hours) -- with idempotent tasks, $12$--$24$-hour
task timeouts, and both engines pinned by the Dockerfiles shipped in
the code appendix (Debian-packaged Stockfish 15.1; Lc0 v0.31.2, BLAS
CPU backend, t1-256x10 distilled network). Everything else -- training,
featurization, and evaluation for both games -- runs on Apple Silicon
(M5 Max, $128$\,GB unified memory, macOS 26.4), using PyTorch 2.13 on
the MPS backend under Python 3.12, with KataGo v1.16.5 for the Go
engine stages (Go appendix).

\section{The Go generalization: data, protocol, and full results}\label{app:go}
The main paper's Beyond Chess: Go section reports the headline
marginals; this appendix carries the full experiment. Everything regenerates from
\texttt{go\_pipeline/reproduce.sh} (cached engine artifacts are public, so
the tables and figures reproduce in minutes without an engine).

\subsection{The port}
The architecture transfers unchanged. The frozen human prior is KataGo's
human-imitation network \texttt{b18c384nbt-humanv0}
\citep{katagohumansl,goprior}, conditioned on player rank (amateurs) or
era (\texttt{proyear}, professionals); it predicts the move a human of
that skill or era would play. The search evidence comes from KataGo's
\texttt{g170 b10c128} network \citep{katagonetworks} run as a search over
each position's candidate moves. The re-ranker is the same residual
design as \textsc{Matilda}: a $0.5$M-parameter network over the prior's
top candidates whose zero-initialized head reproduces the prior exactly
before training. Per-candidate search features are the Go analogue of the
centipawn block: winrate, score lead, lead loss versus the best-scoring
candidate, search rank, and a search-top-choice flag. As in chess, two
models are trained on identical data with the same seed, differing only
in whether the search-feature block is present; their difference -- the
search marginal -- isolates what the search evidence contributes.
Personalization is the same per-player embedding mechanism.

\subsection{Data and split}
\textbf{Amateurs (KGS).} u-go.net game records \citep{ugokgs}, 2005--2015:
ranked games (KGS's rated equivalent) on the full 19$\times$19 board,
even -- neither side receives the handicap stones that compensate a rank
gap -- in two bands: $5$--$6$
dan and $7$ dan$+$. The window ends before the 2016 AlphaGo match
\citep{masteringgo}, so amateur play here predates engine influence.
Players contribute $7$--$8$ training games on average (at most $65$).
\textbf{Professionals.} Official tournament games from kifubara.app
(GoGoD records) \citep{gogod,kifubara}: $5{,}000$ games played in 2025
(AI era) and $5{,}000$ played in 2014--15 (pre-AI) -- two batches roughly
ten years apart, cleanly on either side of the 2016 AlphaGo--Lee
Sedol match. Per-player
temporal selection keeps $120$ players (AI era) and $121$ (pre-AI) with
three training games each ($\sim$$330$ training decisions per player):
$40{,}142$ training and $13{,}625$ evaluation decisions in the AI-era
batch, $40{,}101$ and $13{,}158$ pre-AI. When scoring 2025 games the prior's era conditioning
is capped at its newest profile, \texttt{proyear} 2023 -- a two-year
mismatch that can only understate the prior and is therefore
conservative.

\subsection{Results}
\begin{figure}[htbp]
\centering
\includegraphics[width=0.72\linewidth]{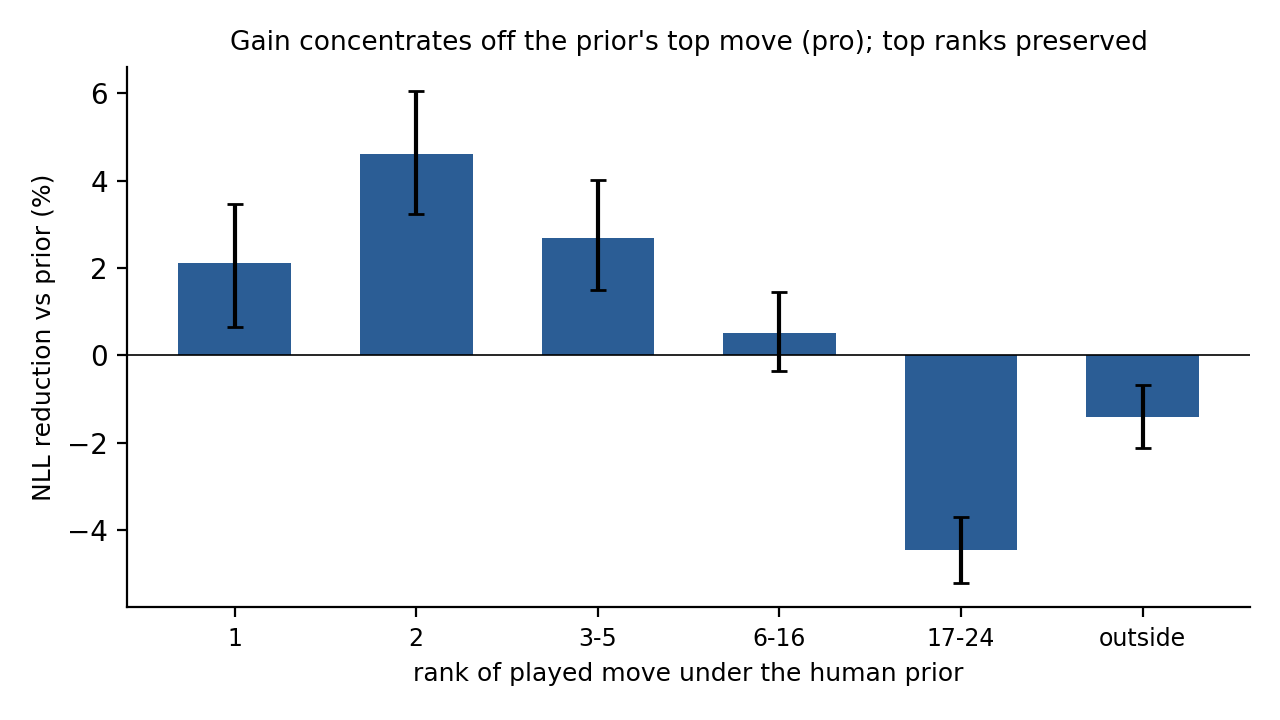}
\caption{Go professionals (AI era): re-ranking pays where the prior's
rank of the played move was wrong -- the chess pattern reproduced.}
\label{fig:gorankcond}
\end{figure}

\begin{figure}[htbp]
\centering
\includegraphics[width=0.8\linewidth]{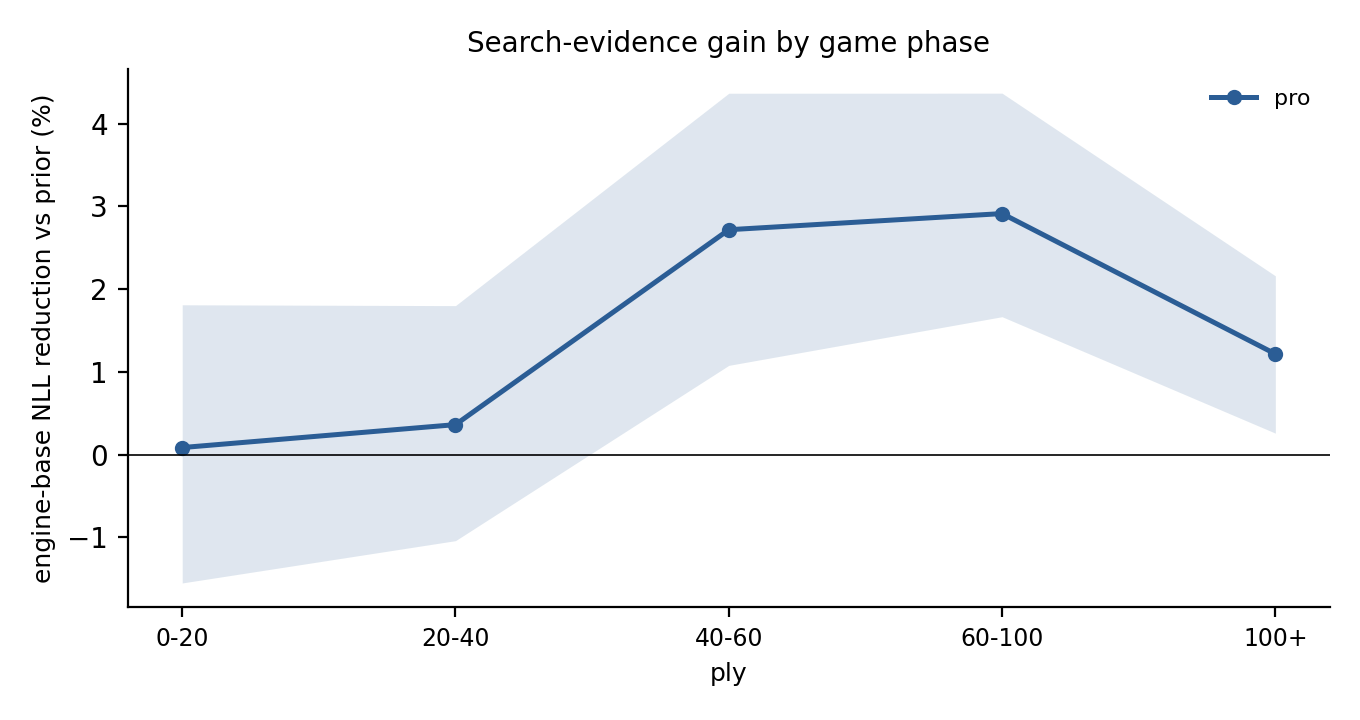}
\caption{Go professionals (AI era): search-evidence gain by game phase.}
\label{fig:goply}
\end{figure}

Table~\ref{tab:gobands} is the full version of the main paper's Go
figure. The with-search model beats the prior everywhere -- but so does
the no-search control; the seed-paired difference is what separates the
populations. The search marginal is $\sim$$0$ for both amateur bands and
clearly positive for both professional eras: $+2.22\%$ [$1.56,2.91$] in
2025 and $+1.82\%$ [$1.17,2.45$] in 2014--15. The era contrast -- the
protocol-matched comparison this experiment was designed around -- is
null: $+0.4$pp with overlapping intervals. Professional convergence on
search-approved moves is not an engine-era phenomenon; what the search
evidence prices is professional skill itself, which the amateur bands,
even at $7$ dan, do not reach.
Figure~\ref{fig:gorankcond} and Figure~\ref{fig:goply} give the same
qualitative breakdowns as the chess appendix: the gain concentrates on
decisions where the prior's ranking was wrong, and grows through the
game rather than living in memorized openings. The payoff concentrating
on a few top candidates mirrors de Groot's account of expert
investigation -- deep verification of only a handful of candidate moves
\citep{degroot1965thought} -- now reproduced in a second game.

\begin{table}[htbp]
\centering
\caption{Go populations: evaluation size, the frozen prior's absolute
performance, the with-search model's gain, and the seed-paired search
marginal ($95\%$ player-stratified bootstrap CIs, 2000 resamples).}
\label{tab:gobands}
{\small\setlength{\tabcolsep}{3pt}
\begin{tabular}{lrrrrr}
\toprule
group & eval $n$ & prior NLL & prior top-1 \% & $+$search vs prior \% & search marginal \% [95\% CI] \\
\midrule
KGS amateur 5--6 dan & 19,983 & 1.651 & 52.3 & +2.66 & -0.12 [-0.26,0.01] \\
KGS amateur 7 dan$+$ & 20,130 & 1.586 & 53.5 & +2.61 & +0.04 [-0.11,0.17] \\
professionals 2014--15 (pre-AI) & 13,158 & 1.583 & 52.1 & +2.56 & +1.82 [1.17,2.45] \\
professionals 2025 (AI era) & 13,625 & 1.447 & 55.5 & +1.87 & +2.22 [1.56,2.91] \\
\bottomrule
\end{tabular}
}
\end{table}

\subsection{Mechanism slices}
\begin{figure}[htbp]
\centering
\includegraphics[width=0.62\linewidth]{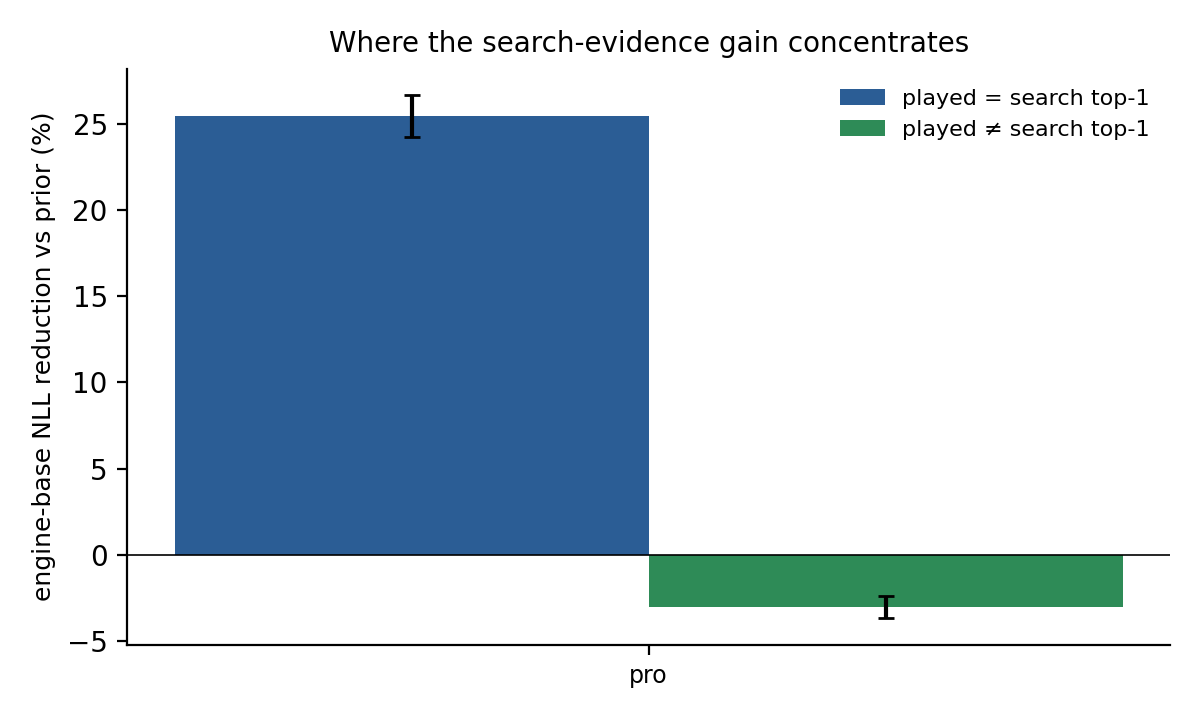}
\caption{Go professionals (AI era): gain split by played-move agreement
with the search's top choice.}
\label{fig:goenginesplit}
\end{figure}

Table~\ref{tab:godecomp} repeats the chess mechanism decomposition
(qualitative-breakdowns appendix) on the AI-era professional rows:
the marginal split by whether the mover played the search's top choice
and whether the prior's own top choice was refuted (here: loses
$\geq$$1$ point of score lead against the best candidate). The chess
pattern reproduces exactly: the whole marginal is carried by rows where
the mover played the search's choice, largest where the prior's choice
was refuted, and negative on deviation rows. Professionals do avoid the
prior's refuted top choice -- $63\%$ of the time versus a $38\%$
baseline -- but they overwhelmingly avoid it by playing the search's
move, so the correction and the alignment bet are the same update
(Figure~\ref{fig:goenginesplit}). The pre-AI decomposition matches
slice for slice ($+28.6\%$ on refuted-agreement rows, negative on both
deviation slices): the mechanism, like the marginal, does not depend on
the era.

\begin{table}[htbp]
\centering
\caption{AI-era professional rows: the search marginal by mechanism
slice (prior's top choice refuted $=$ $\geq$$1$ point of score-lead loss
versus the best candidate). Shares are of the total NLL reduction; the
negative slices make the positive shares exceed one.}
\label{tab:godecomp}
{\small\setlength{\tabcolsep}{3pt}
\begin{tabular}{lrrrrr}
\toprule
 & \multicolumn{3}{c}{2025 (AI era)} & \multicolumn{2}{c}{2014--15 (pre-AI)} \\
\cmidrule(lr){2-4}\cmidrule(lr){5-6}
slice & $n$ & marginal \% & share & $n$ & marginal \% \\
\midrule
played search top-1, prior's choice refuted & 523 & +33.7 & +0.83 & 546 & +28.6 \\
played search top-1, prior's choice sound & 4,331 & +22.6 & +1.19 & 3,775 & +22.7 \\
deviated, prior's choice refuted & 2,930 & -2.2 & -0.27 & 2,907 & -2.1 \\
deviated, prior's choice sound & 5,755 & -3.2 & -0.75 & 5,809 & -2.9 \\
\bottomrule
\end{tabular}
}
\end{table}

\subsection{Annotation protocol difference (KGS vs professionals)}
The two amateur bands and the professional bands were not annotated
identically. The KGS bands used a single $100$-visit search per position
shared across the prior's top-$24$ candidates, so only the candidates the
search chose to visit receive statistics -- $35$--$36\%$ of candidates,
ranked by visit share. The professional bands used per-candidate search:
each of the prior's top-$16$ candidates searched separately at $150$
visits, so every candidate is fully scored and ranked by score lead. The
amateur-versus-professional contrast is therefore not
annotation-matched; the clean, same-protocol comparison in this paper is
\emph{pre-AI versus AI-era professionals}, which differ only in the era
of the games. A second caveat cuts the same way: amateurs who calculate
explicitly only in rare, sharp positions could hide a small real effect
inside the tight $\sim$$0$ intervals -- a marginal confined to a few
percent of decisions would be invisible at this scale. The amateur null
is therefore best read as no measurable marginal under this protocol
and budget, not as proof of absence.

\subsection{Personalization}
Per-player embeddings add nothing for professionals ($-0.12\%$ identity
marginal in 2025, $+0.03\%$ in 2014--15, personal versus generic on the
same rows). This null is
budget-limited -- three training games per player is far below the
chess experiments' per-player history -- so it is not evidence against
personalization as such. The stronger null is KGS, where players
contribute several times more games and the identity marginal is still
$\sim$$0$ ($-0.04\%$ at $7$ dan$+$, $+0.04\%$ at $5$--$6$ dan): under
this protocol, Go style does not measurably separate from Go strength.

\subsection{Game overlap}
The split is per (player, game): a game can contribute one player's
moves to training and the opponent's to evaluation. Re-scoring only
evaluation games never seen in training ($8{,}940$ of $13{,}625$ AI-era
and $9{,}509$ of $13{,}158$ pre-AI professional rows; $9{,}353$ of
$20{,}130$ and $8{,}768$ of $19{,}983$ for the amateur bands) leaves
the conclusions unchanged: the professional marginals hold at $+2.03\%$
[$1.19,2.88$] (2025) and $+1.70\%$ [$0.93,2.46$] (2014--15), and the
amateur marginals stay $\sim$$0$ ($+0.19\%$ [$-0.06,0.44$] and
$+0.09\%$ [$-0.11,0.28$]). The headline comparison is additionally immune by
design: both models of a seed-paired pair share identical training data,
so their difference cancels any overlap effect.

\subsection{Training and hardware}
Every Go model trains for $4$ epochs with AdamW (learning rate
$3{\times}10^{-4}$, weight decay $10^{-4}$); the seed-paired pair
shares seed $0$ for initialization, data order, and schedule. The
personalization stage trains only the embedding table and projection,
$4$ epochs at learning rate $10^{-3}$. All stages run on a single Apple
Silicon machine -- PyTorch on the MPS backend for training and scoring,
KataGo v1.16.5 for the featurization and search stages.

\subsection{Reproduction}
Cached artifacts (models, features, per-candidate annotations, trained
checkpoints, raw professional game files) are publicly readable at
\url{https://storage.googleapis.com/matilda-aaai-submission-2027/go/}.
From the repository, \texttt{bash go\_pipeline/reproduce.sh verify}
recomputes every number and figure in this appendix in about ten
minutes without an engine; \texttt{retrain} and \texttt{scratch} modes
re-train the models or re-run the engine stages. Search annotation
samples a board symmetry per query, so from-scratch numbers match cached
ones to about two decimals; feature extraction is exact.

\end{document}